\documentclass[journal]{IEEEtran}

\usepackage{cite}

\ifCLASSINFOpdf
   \usepackage[pdftex]{graphicx}
   \DeclareGraphicsExtensions{.pdf,.jpeg,.png}
\else
   \usepackage[dvips]{graphicx}
   \DeclareGraphicsExtensions{.eps}
\fi

\usepackage{amsmath}
\usepackage{amsfonts}
\usepackage{bm}
\usepackage{multicol}
\usepackage{multirow}

\usepackage{placeins}

\usepackage{float}

\usepackage{svg}

\usepackage[bookmarks=false]{hyperref}
\hypersetup{hidelinks}
\usepackage[capitalise,nameinlink]{cleveref}

\usepackage{tabularx}
\newcolumntype{Y}{>{\centering\arraybackslash}X}
\newcolumntype{Z}{>{\centering\arraybackslash}p{0.85cm}}
\newcolumntype{K}{>{\centering\arraybackslash}p{0.6cm}}
\newcolumntype{M}[1]{>{\centering\arraybackslash}m{#1}}
\usepackage{booktabs} 

\usepackage{cases}

\usepackage{enumitem}
\usepackage{stfloats}  
\usepackage{slashbox}

\usepackage{xspace}

\usepackage{etoolbox}
\makeatletter
\patchcmd{\@makecaption}
  {\scshape}
  {}
  {}
  {}
\makeatother

\crefname{appsec}{Appendix}{Appendices}

\ifCLASSOPTIONcompsoc
 \usepackage[caption=false,font=normalsize,labelfont=sf,textfont=sf]{subfig}
\else
 \usepackage[caption=false,font=footnotesize]{subfig}
\fi

\usepackage{glossaries}
\newacronym{to}{TO}{Trajectory Optimization}
\newacronym{com}{CoM}{Center of Mass}
\newacronym{mpc}{MPC}{Model Predictive Control}
\newacronym{rhp}{RHP}{Receding Horizon Planning}
\newacronym{ddp}{DDP}{Differential Dynamic Programming}
\newacronym{nlp}{NLP}{Non-Linear Programming}
\newacronym{lg-rhp}{LG-RHP}{Locally-Guided Receding Horizon Planning}
\newacronym{eh}{EH}{Execution Horizon}
\newacronym{ph}{PH}{Prediction Horizon}
\newacronym{nn}{NN}{Neural Network}
\newacronym{mf-rhp}{MF-RHP}{Multi-Fidelity RHP}

\newcommand{\trsp}{{\scriptscriptstyle\top}}

\begin{document}

\title{Online Multi-Contact Receding Horizon Planning via Value Function Approximation}

\author{Jiayi Wang$^{1,2}$, 
        Sanghyun Kim$^{3}$,
        Teguh Santoso Lembono$^{4}$, 
        Wenqian Du$^{1,2}$, 
        Jaehyun Shim$^{1}$, 
        Saeid Samadi$^{1}$,\\
        Ke Wang$^{5}$,
        Vladimir Ivan$^{6}$, 
        Sylvain Calinon$^{4}$, 
        Sethu Vijayakumar$^{1,2}$, 
        and Steve Tonneau$^{1}$
\thanks{$1$~The authors are with the School of Informatics, The University of Edinburgh, United Kingdom.}
\thanks{$2$~The authors are with the Artificial Intelligence Programme, The Alan Turing Institute, United Kingdom.}
\thanks{$3$~The author is with the Department of Mechanical Engineering, Kyung Hee University, South Korea.}
\thanks{$4$~The authors are with the Idiap Research Institute, Switzerland and with the {\'E}cole Polytechnique F{\'e}d{\'e}rale de Lausanne (EPFL), Switzerland.}
\thanks{$5$~The author is with Dyson Limited, United Kingdom.}
\thanks{$6$~The author is with Touchlab Limited, United Kingdom.}
\thanks{e-mail: jiayi.wang@ed.ac.uk}
}

\markboth{IEEE TRANSACTIONS ON ROBOTICS,~Vol.~XX, No.~XX, April~2024}%
{Jiayi Wang \MakeLowercase{\textit{et al.}}: Online Multi-Contact Receding Horizon Planning via Value Function Approximation}

\maketitle

\begin{abstract}
Planning multi-contact motions in a receding horizon fashion requires a value function to guide the planning with respect to the future, e.g., building momentum to traverse large obstacles.
Traditionally, the value function is approximated by computing trajectories in a prediction horizon (never executed) that foresees the future beyond the execution horizon. 
However, given the non-convex dynamics of multi-contact motions, this approach is computationally expensive. 
To enable online \gls{rhp} of multi-contact motions, we find efficient approximations of the value function. 
Specifically, we propose a trajectory-based and a learning-based approach. 
In the former, namely RHP with Multiple Levels of Model Fidelity, we approximate the value function by computing the prediction horizon with a convex relaxed model. 
In the latter, namely Locally-Guided RHP, we learn an oracle to predict local objectives for locomotion tasks, and we use these local objectives to construct local value functions for guiding a short-horizon RHP. 
We evaluate both approaches in simulation by planning centroidal trajectories of a humanoid robot walking on moderate slopes, and on large slopes where the robot cannot maintain static balance. 
Our results show that locally-guided RHP achieves the best computation efficiency (95\%-98.6\% cycles converge online). This computation advantage enables us to demonstrate online receding horizon planning of our real-world humanoid robot Talos walking in dynamic environments that change on-the-fly. 

\end{abstract}

\begin{IEEEkeywords}
Multi-Contact Locomotion, Legged Locomotion, Humanoid Robots,  Optimization and Optimal Control
\end{IEEEkeywords}

\IEEEpeerreviewmaketitle

\section{Introduction}\label{sec:intro}

\begin{figure}[ht]
    \centering
    \def\svgwidth{\columnwidth}
        \input{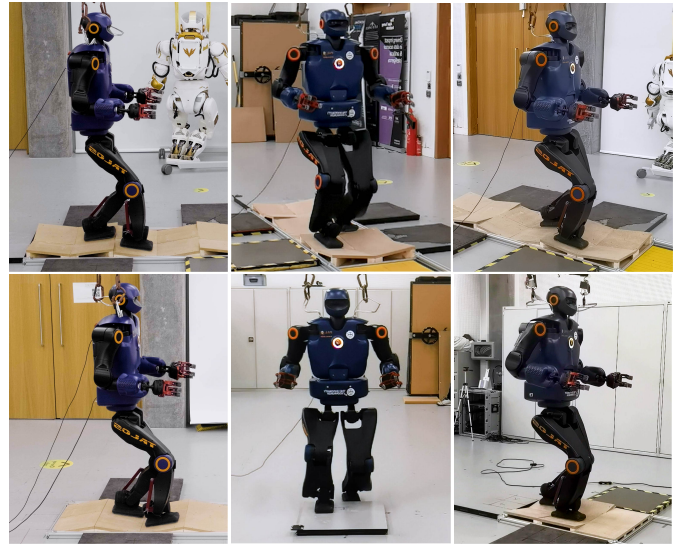}
        \vspace{-7mm}
    \caption{Snapshots of our real-world experiments on Talos. 
    Video is available at \url{https://youtu.be/STBYJl7jvsg}.} 
    \label{fig:teaserA}
\end{figure}

\IEEEPARstart{T}{his} article considers the problem of computing motion plans for legged robots to traverse uneven terrain (non-horizontal surfaces), 
where the planner needs to find a sequence of contacts, along with a feasible state trajectory. 
This problem is known as multi-contact motion planning, which is high-dimensional, nonlinear, and subject to discrete changes of dynamics that arise from breaking and making contacts~\cite{posa2013direct, winkler2018gait, toussaint2018differentiable, stouraitis2020online,deits2014footstep,aceituno2017simultaneous,mordatch2012discovery,kanoulas2017vision,yan2024impact}. 
Given these complexity, traditional robot control methods often plan the multi-contact motions offline, and then track them with a controller~\cite{tonneau2018efficient,winkler2018gait,aceituno2017simultaneous,kalakrishnan2011learning}. 
However, when deploying legged robots in the real world, they can encounter environment changes and state drifts. 
These perturbations can cause the pre-planned motion to become invalid, and online (re)-planning is needed~\cite{tonneau2018efficient, park2015online,melon2021receding,fankhauser2018robust}. 
To facilitate reliable operation in the real world, our long-term objective is to enable legged robots with the capability to online re-plan their motions. 

Towards this end, Receding Horizon Planning (\gls{rhp})~\cite{park2015online,melon2021receding} can be a promising solution. 
The concept of Receding Horizon Planning (\gls{rhp})~\cite{park2015online,melon2021receding} is similar to \gls{mpc}~\cite{crocoddyl20icra,neunert2018whole,di2018dynamic,Ewen2021whole} in that they both aim to constantly update the optimal actions for immediate execution based on the robot state and the environment. 
In \gls{mpc}, the optimal actions correspond to the optimal control commands for tracking a reference trajectory, while in \gls{rhp}, the optimal actions refer to the motion planned for execution. In both \gls{mpc} and \gls{rhp}, the optimal action is computed by solving a finite horizon \gls{to} problem~\cite{betts2010practical}. 

\begin{figure}[t]
    \centering
    \def\svgwidth{\columnwidth}
\begingroup%
  \makeatletter%
  \providecommand\color[2][]{%
    \errmessage{(Inkscape) Color is used for the text in Inkscape, but the package 'color.sty' is not loaded}%
    \renewcommand\color[2][]{}%
  }%
  \providecommand\transparent[1]{%
    \errmessage{(Inkscape) Transparency is used (non-zero) for the text in Inkscape, but the package 'transparent.sty' is not loaded}%
    \renewcommand\transparent[1]{}%
  }%
  \providecommand\rotatebox[2]{#2}%
  \newcommand*\fsize{\dimexpr\f@size pt\relax}%
  \newcommand*\lineheight[1]{\fontsize{\fsize}{#1\fsize}\selectfont}%
  \ifx\svgwidth\undefined%
    \setlength{\unitlength}{212.625bp}%
    \ifx\svgscale\undefined%
      \relax%
    \else%
      \setlength{\unitlength}{\unitlength * \real{\svgscale}}%
    \fi%
  \else%
    \setlength{\unitlength}{\svgwidth}%
  \fi%
  \global\let\svgwidth\undefined%
  \global\let\svgscale\undefined%
  \makeatother%
  \begin{picture}(1,0.50970018)%
    \lineheight{1}%
    \setlength\tabcolsep{0pt}%
    \put(0,0){\includegraphics[width=\unitlength,page=1]{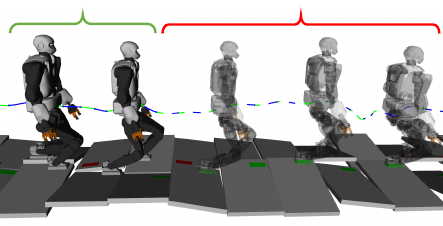}}%
    \put(0.04232804,0.4795358){\color[rgb]{0.00784314,0.00784314,0.01176471}\makebox(0,0)[lt]{\lineheight{1.25}\smash{\begin{tabular}[t]{l}Execution Horizon\end{tabular}}}}%
    \put(0.52347372,0.47892945){\color[rgb]{0.00784314,0.00784314,0.01176471}\makebox(0,0)[lt]{\lineheight{1.25}\smash{\begin{tabular}[t]{l}Prediction Horizon\end{tabular}}}}%
  \end{picture}%
\endgroup%

    \vspace{-7mm}
\caption{In Receding Horizon Planning (RHP), the planning horizon  often consists of two parts: 1) execution horizon which plans the motion for immediate execution, and 2) prediction horizon (not executed) that looks into the future.
The prediction horizon serves as an approximation of the value function, which guides the execution horizon by telling whether the decisions made in the execution horizon can facilitate the completion of the task or not.}
\label{fig:rhp_example}
\end{figure}

To ensure successful multi-contact \gls{rhp} on uneven terrain, it is critical that the motion planned for execution can facilitate future operation. 
For instance, building momentum in advance is often necessary for overcoming large obstacles (slopes and gaps). 
To this end, Bellman suggests to leverage a value function to guide the planning of optimal actions~\cite{bellman1966dynamic}. 
This value function is designed to tell the utility of a certain state with respect to the completion of a given task. 
Nevertheless, for complex dynamical systems, finding an exact model of the value function is challenging, and thus we need approximations. 
A common approach to approximate the value function is to consider a prediction horizon (not executed) that foresees the future beyond the execution horizon (optimal actions to be executed). 
This prediction horizon can be seen as a trajectory-based approximation of the value function---which guides the execution horizon by assessing the feasibility and the anticipated effort required to complete the task from a given state (see an example in \cref{fig:rhp_example}).

Traditionally, Receding Horizon Planning (\gls{rhp}) frameworks often compute the entire horizon with an accurate dynamics model. 
This ensures the execution horizon is always dynamically consistent, while in the meantime allowing the prediction horizon to approximate the value function as accurately as possible. 
However, planning the prediction horizon with an accurate model can result in expensive computation, especially when long planning horizon and complex dynamics need to be considered, i.e., planning multi-contact motions to traverse large slopes. 
In this work, we consider the traditional \gls{rhp} approach as our baseline. 

To accelerate the computation speed, one of the options is to reduce the computation burden required for approximating the value function. 
Following this idea, we propose a trajectory-based approach and a learning-based approach that can improve the computational efficiency for achieving value function approximation.
We compare these two approaches in the context of planning centroidal trajectories of the humanoid robot Talos~\cite{stasse2017talos} walking on uneven terrain. 

More specifically, our trajectory-based approach---which we call Receding Horizon Planning with Multiple Levels of Model Fidelity---follows the traditional formalism that models the value function with the trajectories planned in the prediction horizon. 
However, instead of considering accurate dynamics models, we relax the model accuracy of the prediction horizon.
This allows us to reduce the overall complexity of the \gls{rhp} problem.  
In this article, we explore and compare three candidate multi-fidelity \gls{rhp}s, where the prediction horizon considers different convex relaxations of the centroidal dynamics model (examples shown in \cref{fig:candidates_for_prediction_horizon}).

Alternatively, we can further improve the computation efficiency by approximating the value function with a learned model~\cite{zhong2013value}. 
Nevertheless, learning a value function for multi-contact problem can be challenging. 
The main difficulty is that the value function is defined in a coupled state-environment space, which requires a flexible representation to capture the landscape changes of the value function with respect to different environments~\cite{deits2019lvis}. 
In this article, we circumvent this issue by learning an oracle to predict local objectives (intermediate goal states towards the completion of a given task) based on the current robot state, goal position, and the environment model. 
We then construct local value functions based on these local objectives, and use them to guide a short-horizon \gls{rhp} to plan the execution horizon towards the predicted local objectives. 
We refer to this approach as \gls{lg-rhp}. 
To obtain the oracle, we take a supervised learning approach, where we train the oracle from the dataset offline computed by the traditional \gls{rhp} that computes the entire horizon with an accurate model. 

To evaluate the performance of multi-fidelity \gls{rhp} and locally-guided \gls{rhp}, we consider an online receding horizon planning setting, where we require each cycle to converge within a time budget---the duration of the motion to be executed (execution horizon) for the current cycle. 
From our experiment result, we obtain the following insights. 
First, the result of multi-fidelity \gls{rhp} demonstrates that it is possible to achieve online computation by trading off the model accuracy in the prediction horizon. 
However, this can affect the accuracy of the value function modeled by the prediction horizon. 
As a consequence, our multi-fidelity \gls{rhp} has the risk to arrive at ill-posed states, from which the \gls{to} can fail to converge.
Additionally, we also notice that incorporating angular dynamics in the prediction horizon is critical to the convergence of multi-contact \gls{rhp}. 
On the other hand, as locally-guided \gls{rhp} features a shortened planning horizon, it achieves the highest online convergence rate (95.0\%-98.6\% cycles computes online) compared to the traditional \gls{rhp} (baseline) and multi-fidelity \gls{rhp}. 
Nevertheless, due to the prediction error of the oracle, our locally-guided \gls{rhp} can also arrive at ill-posed states and fail to converge. 
We show that this issue can be mitigated by a data augmentation technique, in which we add datapoints to demonstrate how to recover from the states that cause convergence failures. 

To validate our methods, we verify the dynamic feasibility of the planned trajectories by tracking them with a whole-body inverse dynamics controller~\cite{del2016robustness} in simulation. 
Furthermore, we validate locally-guided \gls{rhp} with real-world experiments, where we demonstrate online receding horizon planning of multi-contact motions on our humanoid robot Talos in dynamically changing environments (see examples in \cref{fig:teaserA} and \cref{fig:changing_env_fig}). The video of the experiments can be found in \url{https://youtu.be/STBYJl7jvsg}.

\subsection{Contributions}
We propose two novel methods that can achieve online Receding Horizon Planning (\gls{rhp}) of centroidal trajectories for multi-contact locomotion. 
The key idea of our methods is to reduce the computation complexity by finding computationally efficient approximations of the value function.
Our contributions are: 
\begin{itemize}
    \item Receding Horizon Planning with Multiple Levels of Model Fidelity, where we approximate the value function by computing trajectories in the prediction horizon while considering convex relaxed models. This allows us to reduce the overall computation complexity of the \gls{to} and facilitates online computation. 
    
    \item Locally-Guided Receding Horizon Planning (LG-RHP), where the value function is approximated with a learned oracle. This oracle is designed to predict local objectives as intermediate goal states for completing a given task, while taking into account the environment model around the robot. We use these local objectives to build local value functions for guiding a short-horizon \gls{to} to plan the execution horizon.

    \item Extensive evaluations and analysis on the computation performance of multi-fidelity \gls{rhp} and locally-guided \gls{rhp}, along with the validation of the dynamic feasibility of the planned trajectories using a whole-body inverse dynamics controller in simulation. 

    \item Real-world experiments on the humanoid robot Talos that demonstrate the effectiveness of our locally-guided \gls{rhp} approach in achieving online multi-contact receding horizon planning on uneven terrain and environments with dynamic changes.    
    
\end{itemize}

\subsection{Comparison with Our Previous Work and Article Outline}
This article is an extension of our earlier conference papers~\cite{wang2021multi} and~\cite{wang2022learning}, where we initially proposed the idea of multi-fidelity \gls{rhp} and the locally-guided \gls{rhp}. 
Compared to our previous work, the novel content of this article includes the following parts. First, we unify the description of the \gls{rhp} problem and the concept of multi-fidelity \gls{rhp} and locally-guided \gls{rhp} under the framework of Bellman's principle of optimality~\cite{bellman1966dynamic}. Second, we conduct a rigorous simulation evaluation on the computation performance of the multi-fidelity \gls{rhp} and the locally-guided \gls{rhp} over a set of multi-contact scenarios. Third, we perform multiple real-world experiments on our humanoid robot Talos showing the efficacy of locally-guided \gls{rhp} in achieving online receding horizon planning. We consider environments that can change dynamically during run-time and challenging uneven terrains.
Lastly, we provide a qualitative analysis on the advantages and disadvantages of multi-fidelity \gls{rhp} and locally-guided \gls{rhp}. 

The rest of the paper is organized as follows. 
\cref{sec:relatedwork} reviews the literature on optimization-based multi-contact locomotion planning, and learning-based methods for accelerating their computation speed. 
\cref{sec:problem_description} describes the \gls{rhp} problem, and introduces the principle of multi-fidelity \gls{rhp} and locally-guided \gls{rhp}. 
\cref{sec:assumptions} lists the assumptions made in our work, and \cref{sec:traditional_to} presents the baseline approach---the traditional \gls{rhp} which computes the entire horizon with an accurate dynamics model. 
\cref{sec:multi_fidelity_rhp} and \cref{sec:locall_guided_rhp} introduce the technical approach of multi-fidelity \gls{rhp} and the locally-guided \gls{rhp}. 
\cref{sec:result} presents our simulation studies, and \cref{sec:real_world_exp} demonstrates the real-world experiment result on our humanoid robot Talos. 
In \cref{sec:discussion}, we discuss the advantages and disadvantages of multi-fidelity \gls{rhp} and locally-guided \gls{rhp}, and we conclude the article in \cref{sec:conclusion}. 

\section{Related work}\label{sec:relatedwork}

\subsection{Multi-Contact Motion Planning via \gls{to}}\label{sec:locomotion_to}
Planning multi-contact motions to traverse challenging terrain necessarily requires the consideration of the whole-body dynamics of the robot. 
This model takes into account the mass and inertia of every link and relates the base and joint accelerations with respect to the joint torques.
In the past, \gls{to}-based methods have demonstrated impressive motions using the whole-body dynamics model~\cite{posa2013direct,tassa2012synthesis,schultz2009modeling,koch2012optimization,erez2012trajectory,koschorreck2012modeling}. 
However, these approaches often struggle to compute online due to the high-dimensionality and non-convexity of the model, unless we predefine the contact timings and locations~\cite{mastalli2022feasibility,meduri2022biconmp}. 

Alternatively, we can plan multi-contact motions with the centroidal model~\cite{orin2013centroidal,carpentier2018multicontact}. 
This model has lower dimensionality since it only considers the dynamics of the total linear and angular momenta expressed at the \gls{com}.
Moreover, approximations are introduced on the robot kinematics and the momentum variation results from the motions of each individual link. 
Although these approximations may cause failures for achieving a corresponding whole-body motion, the centroidal model is getting popular for multi-contact planning due to its reduced dimensionality~\cite{dai2014whole,herzog2016structured,carpentier2018multicontact,wensing2013generation,herzog2015trajectory,carpentier2016versatile,fernbach2020c}. 
Unfortunately, the centroidal model is still non-convex\footnote{The centroidal model is non-convex due to the cross products (bilinear terms) from the angular dynamics.} except when limiting assumptions (pre-defined gait, flat/co-planar surfaces) are made~\cite{kajita2003biped,englsberger2015three,wieber2008viability}. 
Such non-convexity often prevents the \gls{to} to compute online. 

To accelerate the computation, convex approximations of the centroidal model are proposed. 
For instance,~\cite{fernbach2020c,caron2016multi,caron2017make} propose convex inner approximation that searches for a solution within a subset of all possible trajectories. 
Despite their fast computation, convex inner approximation may fail to find a solution due to the reduced search space~\cite{fernbach2020c}. 
Alternatively, ~\cite{ponton2021efficient,dai2016planning,aceituno2018simultaneous} present convex outer approximation that introduces convex relaxations into the centroidal dynamics model. 
Although the model complexity is reduced, convex outer approximation may generate motions that violate the system dynamics and cause tracking failures. 
To address this issue, ~\cite{ponton2021efficient} propose to iteratively tighten the relaxation. However, this requires the design of a customized optimization solver. 

In this article, we introduce multi-fidelity \gls{rhp}, where in a single optimization formulation, we employ an accurate model in the execution horizon and a relaxed model in the prediction horizon.
This formulation is straightforward to implement and can be solved directly with off-the-shelf \gls{nlp} solvers. 
Furthermore, the combination of the accurate model and the relaxed model guarantees the dynamic consistency of the motion to be executed (execution horizon), while in the meantime reduce the overall computation complexity of the \gls{to} problem. 

A similar approach to our multi-fidelity \gls{rhp} method is also introduced in \cite{li2021model}. 
In that work, the authors present a \gls{mpc} framework based on \gls{ddp} that combines whole-body dynamics and a non-convex model with reduced order (single-rigid body model~\cite{winkler2018gait,wensing2013generation}) in a single formulation.
Successful demonstrations of 2D quadrupedal locomotion and humanoid running has been shown on flat surfaces. 
In contrast, our emphasis is \gls{rhp} of centroidal trajectories for a humanoid robot to traverse uneven terrain. 
This problem requires careful selection of contact locations and timings, as well as the modulation of the centroidal momenta.
In this regard, the relaxed model employed in the prediction horizon needs to be carefully designed, as the quality of the model can significantly affect the accuracy of the value function approximated by the prediction horizon.  
Furthermore, instead of searching for non-convex models with reduced order, we focus on finding convex relaxations for the prediction horizon. 

\subsection{Learning to Accelerate Multi-Contact Motion Planning}\label{sec:value_function_learning}
Recently, machine learning techniques have gained popularity for bootstrapping the computation of locomotion planning. 
For instance, \cite{lin2019efficient} proposes to learn the evolution of the centroidal momenta, which can guide an A* planner to generate contact plans. 
Another line of research tries to accelerate the computation speed of \gls{to}.
For example,~\cite{Ewen2021whole,Lembono20ICRA,melon2021receding} propose to learn (near)-optimal solutions to warm-start \gls{to}.
Alternatively,~\cite{zhong2013value, deits2019lvis, parag2022value, viereck2022valuenetqp} propose to shorten the planning horizon with a learned value function model placed as the terminal cost. 
Following this idea, our locally-guided \gls{rhp} focuses on learning a value function model for multi-contact planning. 
However, learning a value function for the multi-contact problem is challenging. 
The main difficulty is that the value function is defined in a coupled state-environment space, which requires a flexible parameterization that can capture the landscape changes of the value function with respect to environment variations~\cite{deits2019lvis}. 
To deal with this issue, we propose to learn an oracle to predict intermediate goal states for completing a given task based on the current state, the final goal, and the environment, and then we construct local value functions based on these intermediate goal states. 

Nevertheless, when predicting sequential actions with a learned model, the prediction accuracy can decrease dramatically once the robot reaches a state that is unexplored in the training dataset. This problem is known as \emph{distribution shift}~\cite{ross2011reduction}, which can be mitigated by data augmentation, i.e. adding demonstrations from the states that either appeared from the roll-out of the learned policy~\cite{ross2011reduction, venkatraman2014data}, or sampled from the expert policy with injected noise~\cite{laskey2017dart}. 
In this work, we present a similar data augmentation strategy which focuses on demonstrating corrective actions from the states that cause convergence failures.

\begin{figure*}[ht!]
    \centering
    \def\svgwidth{\linewidth}
        \input{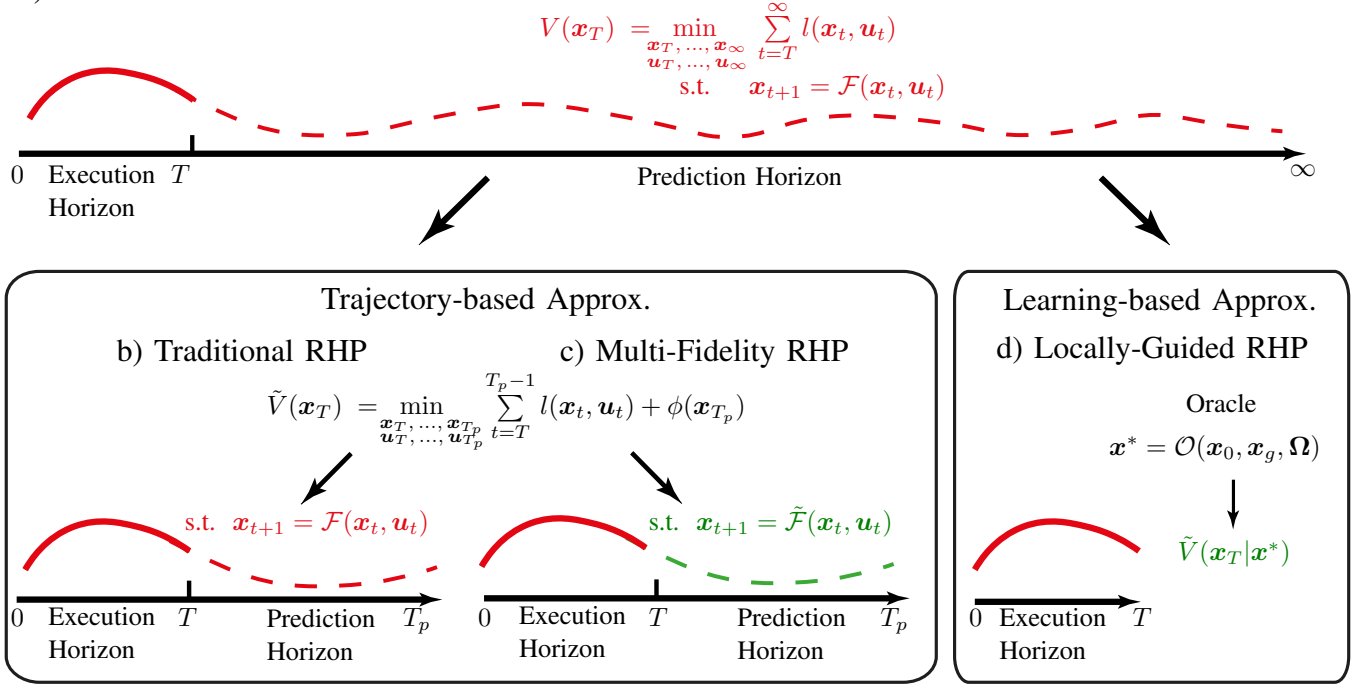}
        \vspace{-7mm}
\caption{a) Infinite-horizon \gls{rhp} problem that models the value function with the prediction horizon of an infinite length; b) Traditional \gls{rhp} approach which approximates the value function by considering a finite-length prediction horizon (from time $T$ to $T_p$). Nevertheless, traditional \gls{rhp} struggles to computes online, as the prediction horizon considers an accurate dynamics model (usually non-convex); c) Multi-fidelity \gls{rhp}, where we improve the computation efficiency by relaxing the model accuracy in the prediction horizon; d) Locally-Guided \gls{rhp} shortens the planning horizon by approximating the value function with a learned model. }
\label{fig:bellman_full}
\end{figure*}

\section{Problem Description}\label{sec:problem_description}
Let us denote by $\bm{x} \in \mathbb{R}^n$ the robot state and $\bm{u} \in \mathbb{R}^m$ the control input. 
In Receding Horizon Planning (\gls{rhp}), each cycle is required to compute a motion plan for immediate execution in the next cycle. 
We define such a motion plan as the composition of a state trajectory $\{\bm{x}_0,\ldots,\bm{x}_{T}\}$ starting from a given initial state $\bm{x}_0$, and a control trajectory $\{\bm{u}_0,\ldots,\bm{u}_{T}\}$.
To compute this motion plan, Receding Horizon Planning (\gls{rhp}) frameworks usually need to solve a \gls{to} problem with the general form complying with the Bellman's equation~\cite{bellman1966dynamic}: 
\begin{IEEEeqnarray}{c'l}\label{eq:bellman_defined_to}
    \IEEEyesnumber
    \IEEEyessubnumber*
    \min_{\substack{\bm{x}_{0},\ldots,\bm{x}_{T}, \\
    \bm{u}_{0},\ldots,\bm{u}_{T}}} & \sum\limits_{t=0}^{T-1}l(\bm{x}_t, \bm{u}_t) + V(\bm{x}_T)
    \label{eq:general_cost_bellman}\\
    \text{s.t.} & \bm{x}_{t+1} = \mathcal{F}(\bm{x}_t,\bm{u}_t), 
    \label{eq:general_dyn_const}
\end{IEEEeqnarray}
where $l(\cdot)$ is the running cost, $V(\cdot)$ is the value function, \eqref{eq:general_dyn_const} is the system dynamics constraint, and $\mathcal{F}(\cdot)$ represents the discrete-time dynamics of the robot. 
As Bellman suggests, the optimal policy---approximated by the motion plan to be executed---should not only minimize its running cost $l$, but also lead to a state $\bm{x}_T$ that optimizes the value function $V(\bm{x}_T)$. 
By definition, the value function is modeled as the optimal cost of an infinite-horizon trajectory starting from $\bm{x}_T$ till the completion of the task, while respecting the system dynamics constraint (\cref{fig:bellman_full}-a):
\begin{IEEEeqnarray}{c'l}\label{eq:value_function_indefinte_horizon}
    \IEEEyesnumber
    \IEEEyessubnumber*
    V(\bm{x}_T) = \min_{\substack{\bm{x}_{T},\ldots,\bm{x}_{\infty}, \\ \bm{u}_{T},\ldots,\bm{u}_{\infty}}} & \sum\limits_{t=T}^{\infty}l(\bm{x}_t, \bm{u}_t)
    \label{eq:general_cost_bellman_full}\\
    \;\;\;\;\;\;\;\;\;\;\;\;\;\;\;\text{s.t.} & \bm{x}_{t+1} = \mathcal{F}(\bm{x}_t,\bm{u}_t).
    \label{eq:general_dyn_const_1}
\end{IEEEeqnarray}

The value function $V(\bm{x}_T)$ reflects the feasibility and the future effort required for accomplishing the given task starting from any state $\bm{x}_T$, 
and provides gradients to direct the optimal policy (motion to be executed) towards a state $\bm{x}_T$ that is favorable for the future. 
However, evaluating the value function with an infinite-horizon trajectory is non-trivial, and hence we need approximations. 

Traditionally, \gls{rhp} frameworks approximate the value function by considering a finite-horizon trajectory starting from the time $T$ to $T_p \ll \infty$: 
\begin{IEEEeqnarray}{c'l}\label{eq:value_function_finite_to}
    \IEEEyesnumber
    \IEEEyessubnumber*
    \tilde{V}(\bm{x}_T) = \min_{\substack{\bm{x}_{T},\ldots,\bm{x}_{T_p}, \\  \bm{u}_{T},\ldots,\bm{u}_{T_p}}} & \sum\limits_{t=T}^{T_p-1}l(\bm{x}_t, \bm{u}_t) + \phi(\bm{x}_{T_p}), 
    \label{eq:general_cost_value_function_finite_to}\\
    \;\;\;\;\;\;\;\;\;\;\;\;\;\;\;\text{s.t.} & \bm{x}_{t+1} = \mathcal{F}(\bm{x}_t,\bm{u}_t),
    \label{eq:general_dyn_const_2}
\end{IEEEeqnarray}
where the optimal cost from the time $T_p$ to infinity is lumped into the terminal cost term $\phi(\bm{x}_{T_p})$. 
By combining \eqref{eq:value_function_finite_to} into \eqref{eq:bellman_defined_to}, we can achieve a \gls{to} problem with an extended planning horizon (\cref{fig:bellman_full}-b): 
\begin{IEEEeqnarray}{c'l}\label{eq:finite_horizon_to}
    \IEEEyesnumber
    \IEEEyessubnumber*
    \min_{\substack{\bm{x}_{0},\ldots,\bm{x}_{T_p}, \\ \bm{u}_{0},\ldots,\bm{u}_{T_p}}} & \underbrace{\sum\limits_{t=0}^{T-1}l(\bm{x}_t, \bm{u}_t)}_\text{Optimal Policy (EH)} + \underbrace{\sum\limits_{t=T}^{T_p-1}l(\bm{x}_t, \bm{u}_t) + \phi(\bm{x}_{T_p})}_\text{Value Function Approximation (PH)}
    \label{eq:general_cost_finite_horizon_to}\\
    \text{s.t.} & \bm{x}_{t+1} = \mathcal{F}(\bm{x}_t,\bm{u}_t),
    \label{eq:general_dyn_constraints_to_complete}
\end{IEEEeqnarray}
where we can split the planning horizon into an \gls{eh} that computes optimal policy (the motion plan to be executed) from the time $0$ to $T$, and a \gls{ph} that approximates the value function by computing trajectories from the time $T$ to $T_p$.  

Although \eqref{eq:finite_horizon_to} has a finite planning horizon, online computation is still challenging for complex dynamical systems such as legged robots. 
The computation complexity mainly comes from the planning of the Prediction Horizon (\gls{ph}) under the consideration of the nonlinear dynamics constraints \eqref{eq:general_dyn_constraints_to_complete}, which increases the dimensionality and non-convexity of an already challenging problem. 

To improve the computation efficiency, a promising direction is to mitigate the computation burden required for value function approximation. 
In this work, we propose two novel methods that can approximate the value function with reduced computation complexity. 

Our first method follows the trajectory-based formalism which approximates the value function by computing a prediction horizon that looks into the future. 
However, instead of considering an accurate system dynamics constraint (usually non-convex) in the prediction horizon, we propose to plan the prediction horizon with a relaxed system dynamics model. 
This gives rise to a novel \gls{to} formulation features a planning horizon with multiple levels of model fidelity (\cref{fig:bellman_full}-c):
\begin{IEEEeqnarray}{c;l}\label{eq:multi_fidelity_rhp}
    \IEEEyesnumber
    \IEEEyessubnumber*
    \hspace{-5mm}    \min_{\substack{\bm{x}_{0},\ldots,\bm{x}_{T_p}, \\ \bm{u}_{0},\ldots,\bm{u}_{T_p}}}
    &  
    \underbrace{\sum\limits_{t=0}^{T-1}l(\bm{x}_t, \bm{u}_t)}_\text{EH (Accurate)} + \underbrace{\sum\limits_{t=T}^{T_p-1}l(\bm{x}_t, \bm{u}_t) + \phi(\bm{x}_{T_p})}_\text{PH (Relaxed)},
    \vspace{2mm}
    \label{eq:general_cost_multi_fidelity_rhp}\\
    \text{s.t.} & \;\;\forall t \in [0,T]: \nonumber \\ 
                & \;\;\;\;\;\; \bm{x}_{t+1} = \mathcal{F}(\bm{x}_t,\bm{u}_t), \;\;\; \text{(Accurate Model)} \label{eq:multi_fidelity_accurate_model} \\
                & \;\;\forall t \in [T,T_p]: \nonumber \\ 
                & \;\;\;\;\;\;\bm{x}_{t+1} = \mathcal{\tilde{F}}(\bm{x}_t,\bm{u}_t), \;\;\;  \text{(Relaxed Model)} \label{eq:multi_fidelity_relaxed_model}
\end{IEEEeqnarray}
where the \gls{eh} remains to compute the optimal polity with the accurate dynamics model \eqref{eq:multi_fidelity_accurate_model}, while the \gls{ph} approximates the value function with the relaxed dynamics model \eqref{eq:multi_fidelity_relaxed_model}. 
We call this approach as Receding Horizon Planning with Multiple Levels of Model Fidelity or \gls{mf-rhp} for short. 
Comparing to the traditional \gls{to} formalism \eqref{eq:finite_horizon_to}, our multi-fidelity \gls{rhp} ensures the Execution Horizon (\gls{eh}) is always dynamically consistent, while in the meantime reducing the overall computation complexity of the \gls{to}. 
In this work, we present and test three candidate multi-fidelity \gls{rhp}s, where the Prediction Horizon (\gls{ph}) considers different convex relaxations of the centroidal dynamics model. 

Alternatively, another option for approximating the value function is to learn a parametric model $\tilde{V}(\bm{x}|\bm{\theta})$ from the past experience\cite{zhong2013value}. 
Given this learned value function model, we can shorten the planning horizon to only cover the Execution Horizon (\gls{eh}):
\begin{IEEEeqnarray}{c'l}\label{eq:to_with_learned_value_function}
    \IEEEyesnumber
    \IEEEyessubnumber*
    \min_{\substack{\bm{x}_{0},\ldots,\bm{x}_{T}, \\ \bm{u}_{0},\ldots,\bm{u}_{T}}} & \underbrace{\sum\limits_{t=0}^{T-1}l(\bm{x}_t, \bm{u}_t)}_\text{Optimal Policy (EH)} + 
    \underbrace{\tilde{V}(\bm{x}_T|\theta),}_\text{Learned Value Function}
    \label{eq:general_cost_to_with_learned_value_function}\\
    \text{s.t.} & \bm{x}_{t+1} = \mathcal{F}(\bm{x}_t,\bm{u}_t).
    \label{eq:general_dyn_constraints_to_complete_finite_horizon_learned_value_function}
\end{IEEEeqnarray}

However, learning a value function for the multi-contact problem can be challenging. 
The difficulty mainly comes from the consideration of the environment model. 
This introduces the challenge of finding a flexible parameterization that can represent the value function in the coupled state-environment space~\cite{deits2019lvis}. 
To tackle this issue, 
we propose to learn an oracle $\mathcal{O}$ that can predict intermediate goal states $\bm{x}^*$ for completing a given task, based on current robot state $\bm{x}_0$, the final goal state $\bm{x}_g$, and the environment model $\bm{\Omega}$:
\begin{IEEEeqnarray}{r'l}\label{eq:general_oralce}
    \IEEEyesnumber
    \bm{x}^* = \mathcal{O}(\bm{x}_0, \bm{x}_g, \bm{\Omega}).
\end{IEEEeqnarray}
We refer to these intermediate goal states as local objectives, and we use them to construct local quadratic value functions: 
\begin{IEEEeqnarray}{r'l}\label{eq:general_local_value_function}
    \IEEEyesnumber
    \tilde{V}(\bm{x}_T|\bm{x}^*) = (\bm{x}_T - \bm{x}^*)^\trsp (\bm{x}_T - \bm{x}^*).
\end{IEEEeqnarray}
We then use these local value functions to guide the short-horizon \gls{to} \eqref{eq:to_with_learned_value_function} to plan the Execution Horizon (\gls{eh}) towards the predicted local objectives $\bm{x}^*$. 
We call this approach as Locally-Guided Receding Horizon Planning (\gls{lg-rhp}) and illustrate the idea in (\cref{fig:bellman_full}-d). Although it is possible to learn to predict the optimal policy of the Execution Horizon (\gls{eh}) directly from the past experiences, the learning error may lead to trajectories that violate system dynamics and cause tracking failures. Hence, in this work we decide to compute the Execution Horizon (\gls{eh}) using \gls{to}, which guarantees the dynamic feasibility of the motion.
Next, from \cref{sec:assumptions} to \cref{sec:locall_guided_rhp}, we present the technical details of our methods in the context of multi-contact motion planning. 

\section{Assumptions}
\label{sec:assumptions}

We make following assumptions in our work: 

\begin{itemize}
    \item We focus on planning centroidal trajectories of a humanoid robot walking on uneven terrain. We define each step contains three phases: pre-swing (double support), swing (single support), and post-landing (double support). This gives rise to a multi-phase \gls{to} formulation, where in each phase the dynamics and kinematics constraints are characterized by the contact configuration of that phase. 
    
    \item We define the Execution Horizon (\gls{eh}) always covers the motion plan for making a single step (the first three phases), while the Prediction Horizon (\gls{ph}) can plan ahead for multiple steps. The oracle is designed to predict the local objective for making one step. 
    
    \item We model the robot feet as rectangular patches. As commonly done, we model each vertex of the rectangle as a contact point. 
    
    \item We approximate the kinematics constraints of the \gls{com} and the relative positions of the contacts as convex polytopes. 
    To generate these polytopes, we firstly offline sample a large amount of robot configurations, from which we can extract \gls{com} positions and foot locations in a given end-effector frame. Then, we compute the polytopes as the convex hull of these \gls{com} positions and foot locations~\cite{tonneau20182pac}. 
    
    \item We model the environment as a set of rectangular contact surfaces. We predefine the sequence of these contact surfaces in which the swing foot will land upon, while we optimize the contact locations (within the surfaces) and the contact timings. 
    
    \item The swing foot trajectory is interpolated after we compute the centroidal motion plan. This is achieved by connecting the planned contact locations with a spline. 

\end{itemize}

\section{Traditional \gls{rhp} Formulation for Multi-Contact Motion Planning}
\label{sec:traditional_to}

In this section, we present the traditional \gls{rhp} approach for planning centroidal trajectories of a humanoid robot walking on uneven terrain. 
This traditional \gls{rhp} is considered as the baseline of our work. 

We describe the \gls{rhp} problem as follows. In each planning cycle, given a finite planning horizon of $n$ steps, an initial robot state $\bm{x}_{init}$, a final goal state $\bm{x}_g$, and a sequence of contact surfaces $\{\mathcal{S}_1,\ldots,\mathcal{S}_n\}$ that the robot will step upon, the \gls{rhp} framework aims to compute a multi-phase motion plan consists of a state trajectory $\bm{\mathcal{X}}$, a control trajectory $\bm{\mathcal{U}}$, a sequence of contact locations $\bm{\mathcal{P}}$ and a list of phase switching timings $\bm{\mathcal{T}}$. 
We elaborate the definition of these decision variables as follows: 

\begin{itemize}
    \item $\bm{\mathcal{X}} = \{{X}^{1},\ldots,{X}^{N_{ph}}\}$: state trajectory ${X}^{q}$ of all phases $q \in \{1,\ldots,N_{ph}\}$. 
    In each phase, we discretize the state trajectory into $N_k$ knots: ${X}^{q} = \{\bm{x}^q_1,\ldots,\bm{x}^q_{N_k}\}$. 
    We denote the state vector as $\bm{x} = [\bm{c}^\trsp, \dot{\bm{c}}^\trsp, \bm{L}^\trsp]^\trsp$, where $\bm{c} \in \mathbb{R}^3$ is the \gls{com} position,  $\dot{\bm{c}} \in \mathbb{R}^3$ is the \gls{com} velocity, $\bm{L} \in \mathbb{R}^3$ is the total angular momentum expressed at the \gls{com}. 

    \item $\bm{\mathcal{U}} = \{{U}^{1},\ldots,{U}^{N_{ph}}\}$: control input trajectory of all phases. Same as the state trajectory, we discretize each phase of the control trajectory into $N_k$ knots: ${U}^{q} = \{\bm{u}^q_1,\ldots,\bm{u}^q_{N_k}\}$. 
    The control input vector is defined as $\bm{u} = [\bm{f}_1^\trsp,\ldots,\bm{f}^\trsp_{N_c}]^\trsp$ which collects the contact force $\bm{f}_c \in \mathbb{R}^3$ of all contact points $c \in \{1,\ldots,N_c\}$.
    
    \item $\bm{\mathcal{P}} = \{\bm{p}^{1},\ldots,\bm{p}^{n}\}$: a sequence of footstep locations (center of the foot), where $\bm{p}^i \in \mathbb{R}^3$ denotes the contact location of the $i$-th step. The orientation of each footstep is defined as a constant, where the roll and the pitch are given by the orientation of the corresponding contact surface $\mathcal{S}_i$, and the yaw is set to zero degrees. 

    \item $\bm{\mathcal{T}} = \{t^{1},\ldots,t^{N_{ph}}\}$: a list of phase switching timings, where $t^{q}$ indicates the timing when the motion plan switches from phase $q$ to phase $q+1$. Based on these phase switching timings, we can define the time step of each phase $q$ as $\tau^{q} = (t^{q}-t^{q-1})/N_k$.
    
\end{itemize}

To compute the motion plan, the traditional \gls{rhp} usually solves a \gls{to} problem given by: 
\begin{IEEEeqnarray}{r'l}\label{eq:general_formulation}
    \IEEEyesnumber
    \IEEEyessubnumber*
    \min_{\substack{\bm{\mathcal{{X}}}, \bm{\mathcal{{U}}}, \bm{\mathcal{T}}, \bm{\mathcal{P}}}} &  \sum\limits_{q=1}^{N_{ph}}J^{q}({X}^{q}, {U}^{q}) + \phi(\bm{x}_T)
    \label{eq:general_cost_traditional_rhp}\\
    \text{s.t.} & \bm{x}_0 = \bm{x}_{init}\label{eq:initial_condition}\\
                & 0 \leq {t}^{1} \leq \cdots \leq {t}^{N_{ph}} \leq T_{max}\label{eq:phase_switching_time}\\
                & \forall i \in \{1,\ldots,n\} \text{:}\nonumber\\
                & \>\>\>\>\>\> \bm{p}^{i}  \in \mathcal{S}_i\label{eq:contact_location}\\
                & \>\>\>\>\>\> \bm{p}^{i} \in \mathcal{Z}_{i}\label{eq:relative_kinematics}\\
                & \forall q \in \{1,\ldots,N_{ph}\}, \forall k \in \{1,\ldots,N_k\} \text{:} \nonumber\\
                & \>\>\>\>\>\> \bm{c}_k^{q} \in \mathcal{K}^q_l, \>\>\> \forall l \in \mathcal{L}_{cnt}^q\label{eq:com_kinematics}\\
                & \>\>\>\>\>\> \bm{x}^{q}_{k+1} = \mathcal{F}^q(\bm{x}^{q}_k,\bm{u}^{q}_k).\label{eq:dynamics}
\end{IEEEeqnarray}

The cost function of \eqref{eq:general_formulation} includes the running cost $J^{q}$ of each phase and the terminal cost $\phi(\bm{x}_T)$. 
We define the running cost $J^{q} = \sum_{k = 1}^{N_k} \tau^{q}( {{{\ddot{\bm{c}}}^{q}_k}} {}^\trsp{{\ddot{\bm{c}}}^{q}_k} + {\bm{L}^{q}_k}^\trsp \bm{L}^{q}_k)$, which encourages the \gls{to} to generate smooth trajectories by penalizing large \gls{com} accelerations and angular momentum.
The terminal cost is defined as $\phi(\bm{x}_T) = (\bm{x}_T - \bm{x}_g)^\trsp(\bm{x}_T - \bm{x}_g)$, which attracts the terminal state $\bm{x}_T$ to approach the final goal state $\bm{x}_g$. 

To ensure the motion is dynamically consistent, we introduce constraints \eqref{eq:initial_condition}--\eqref{eq:dynamics} described as follows: 
\begin{itemize}
    \item \eqref{eq:initial_condition} enforces the state trajectory to start from the given initial state $\bm{x}_{init}$. 
     
    \item \eqref{eq:phase_switching_time} guarantees the phase switching timings $t^{q}$ to increase monotonically and bounds the maximum motion duration $t^{N_{ph}}$ by $T_\text{max}$. 
    
    \item \eqref{eq:contact_location} restricts each contact location $\bm{p}^{i}$ to stay on the pre-assigned contact surface $\mathcal{S}_i=\{\bm{p} \in \mathbb{R}^3, \bm{d}^T_i \bm{p} = e_i, S_i \bm{p} \leq s_i\}$.
    The equality defines the plane containing the surface, where the surface normal is given by $\bm{d}_i \in \mathbb{R}^3$ and $e_i \in \mathbb{R}$. The inequalities bound the surface by $h$ half-spaces, specified by the constant matrix $S_i \in \mathbb{R}^{h \times 3}$ and the constant vector $s_i \in \mathbb{R}^h$.
    
    \item \eqref{eq:relative_kinematics} implements the relative reachability constraint of the foot steps, where each contact location $\bm{p}^i$ is limited by a reachable workspace  $\mathcal{Z}_{i}$ with respect to the 6-D pose of the previous footstep $\bm{p}^{i-1}$. 
    We represent the reachable workspace as a convex polytope $\mathcal{Z}_{i}: \{\bm{p}^{i} \in \mathbb{R}^3, \bm{Z}_i(\bm{p}^{i} - \bm{p}^{i-1}) \leq \bm{z}_i\}$, along with the orientation aligns to the posture of the previous footstep~\cite{tonneau20182pac,song2021solving}.
    
    \item \eqref{eq:com_kinematics} is the \gls{com} reachability constraint. In each phase $q$, the \gls{com} position at $k$-th knot is restricted to stay within the reachable space $\mathcal{K}_l^q$ established by each foot $l$ in active contact. Similarly, we approximate the reachable space as a convex polytope $\mathcal{K}_l^q: \{\bm{c}^q_k \in \mathbb{R}^3, \bm{K}_l^q(\bm{c} - \bm{p}^q_l) \leq \bm{k}_l^q\}$, where $\bm{p}^q_l \in \mathbb{R}^3$ is the location of the active contact $l$ in phase $q$. The orientation of these polytopes are also aligned to pose of the active contacts~\cite{tonneau20182pac,song2021solving}. 

    \item \eqref{eq:dynamics} imposes the system dynamics constraint. We approximate the integrals by the forward Euler integration scheme, and we consider the centroidal dynamics model~\cite{orin2013centroidal,fernbach2020c}:
    \begin{IEEEeqnarray}[]{cCl}\label{eq:centroidal_dynamics}
        \IEEEyesnumber
        \IEEEyessubnumber*
        \hspace{-1mm} 
        \bm{c}^{q}_{k+1} & =  & \bm{c}^{q}_{k} + \tau^{q}\bm{\dot{c}}^{q}_{k}, 
        \label{eq:com_integration}
        \\
        \bm{\dot{c}}^{q}_{k+1} & = & \bm{\dot{c}}^{q}_{k} + \tau^{q} \bigg(\frac{1}{m}\sum_{c \in \mathcal{C}^{q}}^{} \bm{f}^{q}_{c,k} - \bm{g}\bigg),
        \label{eq:comdot_integration}
        \\
        \bm{L}^{q}_{k+1} & = & \bm{L}^{q}_{k} + 
        \tau^{q}\sum_{c \in \mathcal{C}^{q}}^{} (\bm{p}_c - \bm{c}^{q}_k) \times \bm{f}^{q}_{c,k},
        \label{eq:angular_dynamics}
    \end{IEEEeqnarray}
    where $m$ is the total mass of the robot, $\bm{g}$ is the gravitational acceleration, $\mathcal{C}^q$ is the set that collects the indices of the contact points in active contact with the ground in phase $q$, i.e., the vertices of the stance foot, and $\bm{p}_c$ is the location of each active contact point $c \in \mathcal{C}^q$.
    We constrain the contact force associated to each active contact point by the linearized friction cone $-\mu \bm{f}_{c}^{\hat{n}} \leq \bm{f}_{c}^{{\hat{t}_1},{\hat{t}_2}} \leq \mu \bm{f}_{c}^{\hat{n}}$, where $\mu$ is the friction coefficient, $\bm{f}_{c}^{\hat{n}}$ and $\bm{f}_{c}^{{\hat{t}_1},{\hat{t}_2}}$ are the normal and tangential components of the contact force, respectively. 
\end{itemize}

As discussed in \cref{sec:problem_description}, in addition to the Execution Horizon (\gls{eh}), traditional \gls{rhp} often requires the consideration of a Prediction Horizon (\gls{ph}) that acts as a trajectory-based approximation of the value function. 
However, traditional \gls{rhp} often plans the entire horizon with an accurate dynamics model, e.g., the nonlinear centroidal dynamics model \eqref{eq:centroidal_dynamics}. This can significantly increase the dimensionality and non-convexity of the \gls{to}, which hinders online computation. 
To achieve online multi-contact \gls{rhp}, in the following sections, we present our methods to simplify the computation for value function approximation. 

\begin{figure}[t]
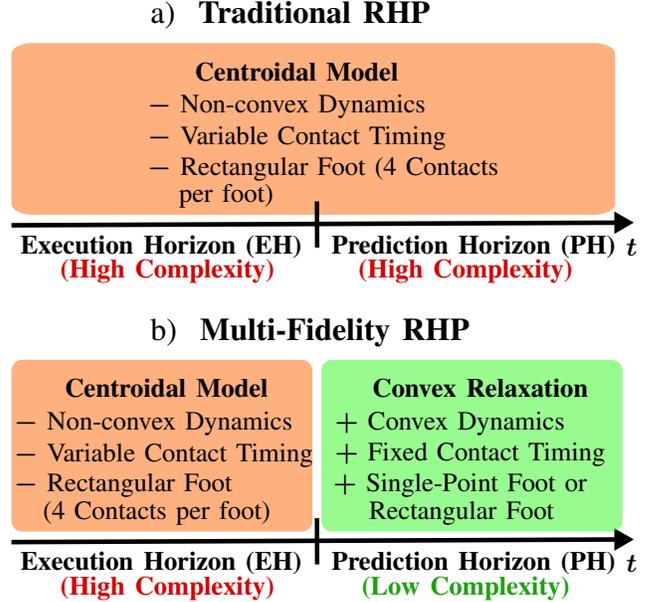

    \captionsetup[subfigure]{labelformat=empty}
    \centering
    
    \subfloat[][]{
    \def\svgwidth{0.95\columnwidth}
    \input{./figures/single_fidelity.tex}
    \label{fig:single_fidelity_diagram}}\\
    
    \vspace{-8mm}
    
    \subfloat[][]{
    \def\svgwidth{0.95\columnwidth}
    \input{./figures/multi_fidelity.tex}
    \label{fig:multi_fidelity}}
    \vspace{-6mm}
    
    \caption{Complexity comparison between traditional \gls{rhp} and our multi-fidelity \gls{rhp}. We use orange to denote higher computation complexity, while green means lower computation complexity. Our multi-fidelity \gls{rhp} formulation has reduced complexity due to the introduction of convex relaxations in the prediction horizon.}
    \label{fig:config_of_rhps}
\end{figure}

\section{\gls{rhp} with Multiple Levels of Model Fidelity}
\label{sec:multi_fidelity_rhp}

In this section, we introduce Receding Horizon Planning (\gls{rhp}) with Multiple Levels of Model Fidelity. 
Unlike the traditional \gls{rhp} (\cref{fig:single_fidelity_diagram}) which computes the entire horizon with an accurate dynamics model, our multi-fidelity \gls{rhp} (\cref{fig:multi_fidelity}) employs a convex relaxation of the centroidal dynamics model in the Prediction Horizon (\gls{ph}) for value function approximation. This simplifies the overall computation complexity of the \gls{to}.
Next, we present three candidate multi-fidelity \gls{rhp}s with different convex relaxations.

\subsection{Candidate 1: Linear \gls{com} Dynamics}
In our first candidate, the \gls{ph} only considers the linear \gls{com} dynamics defined by~\eqref{eq:com_integration}--\eqref{eq:comdot_integration}. 
This allows us to remove the non-convexity introduced by the angular dynamics \eqref{eq:angular_dynamics}. 
As a result, in the \gls{ph}, the state vector reduces to $\bm{x} = [\bm{c}^\trsp, \dot{\bm{c}}^\trsp]^\trsp$ and the running cost becomes $J^{q} = \sum_{k = 1}^{N_k} \tau^{q}( {{{\ddot{\bm{c}}}^{q}_k}} {}^\trsp{{\ddot{\bm{c}}}^{q}_k})$ which only penalizes the \gls{com} accelerations. 
However, due to the contact timing optimization (modulated by the phase switching timings  $t^{q}$), the linear \gls{com} dynamics is still non-convex. 
To eliminate this non-convexity, we choose to fix the phase switching timings $t^{q}, \forall q \in [4, N_{ph}]$ in the \gls{ph}. 

\subsection{Candidate 2: Convex Angular Dynamics with Rectangular Contacts}
For our second candidate, in addition to the linear \gls{com} dynamics, we also consider a convex outer approximation of the angular dynamics \eqref{eq:angular_dynamics} in the \gls{ph}.
This convex approximation is based on the method described in~\cite{ponton2021efficient}. 
For completeness, we briefly introduce the formulation. 

In the angular momentum dynamics \eqref{eq:angular_dynamics}, the non-convexity mainly comes from the bilinear terms result from the cross product between the lever arm $(\bm{p}_c - \bm{c})$ and the contact force vector $\bm{f}_c$. 
According to the principle described in~\cite{ponton2021efficient}, we can approximate each bilinear term $\alpha\beta$ as the difference between two bounded quadratic terms $\psi^{+} \in \mathbb{R}$ and $\psi^{-} \in \mathbb{R}$, along with two convex trust-region constraints:
\begin{IEEEeqnarray}[]{cCl}\label{eq:ponton}
    \IEEEyesnumber
    \IEEEyessubnumber*
    \hspace{-1mm} 
    \alpha\beta & = &\frac{1}{4}(\psi^{+} - \psi^{-}), \\
    \psi^{+} & \geq & (\alpha + \beta)^2, \\
    \psi^{-} & \geq & (\alpha - \beta)^2.
\end{IEEEeqnarray}
Furthermore, to retain a low-dimensional model with the state vector of $\bm{x} = [\bm{c}^\trsp, \dot{\bm{c}}^\trsp]^\trsp$, we avoid the explicit modeling of the angular momentum $\bm{L}$. 
Instead, we penalize the $\psi^{+}$ and $\psi^{-}$ in the running cost as a proxy to minimize the angular momentum rate, along with the \gls{com} acceleration. 
Lastly, same as the first candidate, we fix the phase switching timings in the \gls{ph}. 
Compared to the centroidal dynamics model, our second candidate model has increased dimensionality due to the introduction of the auxiliary variables $\psi^{+}$ and $\psi^{-}$ for approximating the angular dynamics. 
Nevertheless, this also allows our second candidate model to be fully convex.  

\begin{figure}[t]
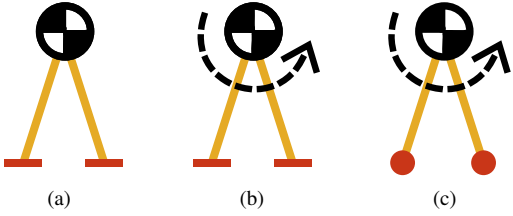

\centering
\subfloat[][]{
\def\svgwidth{0.2\columnwidth}
\input{./figures/CoM_Dynamics.tex}
\label{fig:single_fidelity_com_dynamics}}
\hspace{3mm}
\subfloat[][]{
\def\svgwidth{0.2\columnwidth}
\input{./figures/rectangle_relaxation.tex}
\label{fig:single_fidelity_rectangle_relaxation}}
\hspace{3mm}
\subfloat[][]{
\def\svgwidth{0.2\columnwidth}
\input{./figures/PointFoot.tex}
\label{fig:single_fidelity_point_foot}}

\caption{Schematics of the models used in the Prediction Horizon (PH): a) linear \gls{com} dynamics (Candidate 1); b) convex relaxation of angular momentum rate dynamics (dashed arrow) with rectangular contacts (Candidate 2); c) convex relaxation of angular momentum rate dynamics (dashed arrow) with point contacts (Candidate 3).}
\label{fig:candidates_for_prediction_horizon}
\end{figure}

\subsection{Candidate 3: Convex Angular Dynamics with Point Contacts}
To reduce the dimensionality of the convex relaxation of the angular dynamics, we propose our third candidate model in which we switch the rectangular foot to the point foot and apply the same modeling as described in the second candidate.
As a consequence, the control input reduces to $\bm{u} = [\bm{f}_L^\trsp,\bm{f}_R^\trsp]^\trsp$ where $\bm{f}_L \in \mathbb{R}^3$ and $\bm{f}_R \in \mathbb{R}^3$ refers to the contact force vector of the left and right foot, respectively. 
This reduces the number of auxiliary variables ($\psi^{+}$ and $\psi^{-}$) as well as the associated trust region constraints introduced by \eqref{eq:ponton}.

To provide an intuition of the computation complexity of these candidate models, we illustrate their schematics in ~\cref{fig:candidates_for_prediction_horizon}, and compare their model complexity in terms of dimensionality, number of non-convex and convex constraints in~\cref{tab:model_complexity}.

\begin{table}[t]
  \centering
  \caption{Knot-wise model complexity of the centroidal dynamics model and the three convex relaxations.}
  \label{tab:model_complexity}
  \begin{tabularx}{\linewidth}{M{0.36\linewidth} M{0.12\linewidth} M{0.19\linewidth} M{0.17\linewidth}}
    \toprule
    \backslashbox{Model}{No. of} & Decision variables  & Non-convex Constraints & Convex Constraints \\ 
    \midrule
    Centroidal Dynamics   & 36   & 12   & 0                  \\ 
    Convex (CoM only) & 18 & 0 & 6 \\
    Convex (Rectangular Foot) & 78   & 0   & 48   \\ 
    Convex (Point Foot) & 12   & 0   & 12   \\ 
    \bottomrule
  \end{tabularx}
\end{table}

\section{Locally-Guided Receding Horizon Planning}
\label{sec:locall_guided_rhp}

In this section, we present locally-guided \gls{rhp} (\gls{lg-rhp}) which approximate the value function with a learned model. 
The core idea of the our approach is to learn an oracle that can predict local objectives for completing a given task based on the initial robot state, the final goal and the environment model. 
These local objectives are then used for constructing local value functions that guide the planing of the Execution Horizon (\gls{eh}). 
Next, in \cref{sec:oracle_function}, we present the modeling of the oracle in the context of multi-contact planning. 
Then, we describe the interface to the short-horizon \gls{to} in \cref{sec:interface_to_to_oracle}.

\subsection{Oracle Modelling for Multi-Contact Planning}
\label{sec:oracle_function}

\begin{figure}[t]
    \centering
    \def\svgwidth{\columnwidth}
        \input{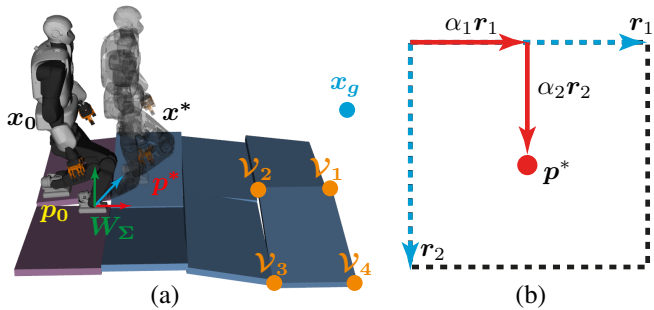}
        \vspace{-7mm}
\caption{(a) Definition of the oracle variables. $\bm{x}_0$ is the initial \gls{com} state, $\bm{p}_0$ is the initial swing-foot position, $\bm{x_g}$ is the final goal state, $\bm{x}^*$ and $\bm{p}^*$ are the predicted target \gls{com} state and the target contact location of the swing foot for making one step, respectively. The purple patches are the initial contact surfaces, while the blue patches are the contact surfaces for future steps. We model each contact surface with its four vertices $\mathcal{V}_i$. We define all spatial terms in the contact foot frame $\bm{W}_{\Sigma}$ established at the stationary foot (non-swing foot), except for the final goal state $\bm{x_g}$ is fixed in the world frame. 
(b) The target contact location $\bm{p}^*$ is represented as the vector sum of $\alpha_1\bm{r}_1$ and $\alpha_2\bm{r}_2$, which scale along the borders of the contact surface $\bm{r}_1, \bm{r}_2 \in \mathbb{R}^3$, with the proportion defined by $\alpha_1, \alpha_2 \in [0,1]$. }
\label{fig:oracle_variables}
\end{figure}

In this section, we firstly describe the oracle formulation in the context of multi-contact planning and introduce the associate variable definitions (see \cref{fig:oracle_variables}-a).
Following the idea in \cref{sec:problem_description}, we define the oracle $\mathcal{O}$ as:
\begin{equation}\label{eq:oracle_multi_contact}
  \begin{IEEEeqnarraybox}[\IEEEeqnarraystrutmode\IEEEeqnarraystrutsizeadd{2pt}{2pt}][c]{rCl}
    \bm{x}^*, \bm{p}^*, \bm{\mathcal{T}}^* = \mathcal{O}(\delta_{l/r}, \bm{x}_0, \bm{p}_0, \bm{\Omega}, \bm{x}_g).
  \end{IEEEeqnarraybox}
\end{equation}

The oracle is designed to predict a goal configuration for making a step, which includes:
\begin{itemize}
    \item $\bm{x}^*$: the target \gls{com} state. 
    \item $\bm{p}^*$: the target contact location for the swing foot to reach. 
    \item $\bm{\mathcal{T}}^* = \{\tilde{t}^1, \tilde{t}^2, \tilde{t}^3\}$: the target phase switching timings for the three phases that compose of the step. 
\end{itemize}

To make the prediction, the oracle takes into account the following inputs: 
\begin{itemize}
    \item $\delta_{l/r} \in \{L,R\}$: the swing foot indicator telling which foot (left/right) is going to re-position its location. 
    \item $\bm{x}_0$: the initial \gls{com} state. 
    \item $\bm{p}_0$: the initial contact location of the swing foot.
    \item $\bm{\Omega} = \{\mathcal{S}_{l0}, \mathcal{S}_{r0}, \mathcal{S}_{1},\ldots,\mathcal{S}_{n}\}$: a local preview of the environment model. 
    We define $\mathcal{S}_{l0}$, $\mathcal{S}_{r0}$ as the contact surfaces that the left and right feet initially stand upon, $\mathcal{S}_{1},\ldots,\mathcal{S}_{n}$ as the contact surfaces that future $n$ steps will land on. Each contact surface is represented by its four vertices: $\mathcal{S} = \{\mathcal{V}_{1},\ldots,\mathcal{V}_{4}\}$, where $\mathcal{V}_i \in \mathbb{R}^3, i \in \{1,\ldots,4\}$ is the 3-D location of the $i$-th vertex. 
    \item $\bm{x}_g$: the final goal state placed in front of the robot (fixed in the world frame). This final goal encourages the robot to continuously move forward.
\end{itemize}

As illustrated in \cref{fig:oracle_variables}-a, we define the spatial quantities such as the \gls{com} states, contact locations and the environment model in the so-called contact-foot frame $\bm{W}_{\Sigma}$. 
This frame locates at the position of the stationary foot (the non-swing foot), while having the same orientation with respect to the surface in contact. 

Furthermore, we introduce an on-surface parameterization for the target contact location (\cref{fig:oracle_variables}-b), which transforms the 3-D contact location as the sum of two vectors $\bm{p}^* = \alpha_1\bm{r_1}+\alpha_2\bm{r_2}$, scaling along the surface borders $\bm{r}_1, \bm{r}_2 \in \mathbb{R}^3$, and we predict the scaling factors $\alpha_1, \alpha_2 \in [0,1]$. 

We model the oracle with a \gls{nn} model. 
Considering the oracle is involved in an online \gls{rhp} loop, it is essential that the \gls{nn} can ensure fast computation. 
To this end, we employ a compact \gls{nn} model with 4 hidden layers, where each layer contains 256 neurons with ReLu activation functions. 
We find that the evaluation of this \gls{nn} only takes 1ms. 
Previously, such compact \gls{nn} models have been proven effective for similar tasks, i.e., predicting the cost and dynamics feasibility for reaching to a given state~\cite{lin2019efficient}, and our experiment result (\cref{sec:result}) also certifies that our \gls{nn} model is flexible enough to encode the oracle. 
Additionally, we also find that increasing the number of hidden layers and neurons does not bring improvements on the prediction accuracy. 
We implement the \gls{nn} with the Tensorflow framework~\cite{abadi2016tensorflow}, and both the training and the prediction are achieved with the CPU mode.

\begin{figure}[t]
    \centering
    \def\svgwidth{\columnwidth}
        \input{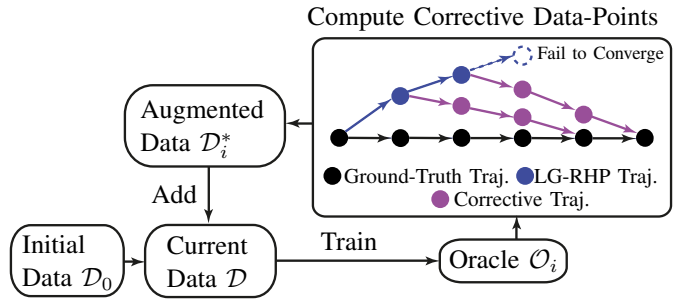}
        \vspace{-6mm}
\caption{Procedure of the incremental training scheme. In each training iteration $i$, we train an oracle $\mathcal{O}_i$ based on the data-set $\mathcal{D}$ that aggregates the initial dataset $\mathcal{D}_0$ and the augmented datasets $\mathcal{D}^*_i$. 
The augmented dataset adds recovery actions (purple nodes) computed by the long-horizon \gls{rhp}. They start from the diverged states (blue nodes) that are 1 to 3 cycles prior to when \gls{lg-rhp} fails to converge (the dashed blue node), until the cycle aligns with the ground-truth trajectory (black nodes). In these trajectories, each node refers to the state reached by the execution horizon (making one step).}
\label{fig:data_augmentation}
\end{figure}

To train the oracle, we employ an incremental training scheme. 
The key idea of our approach is to improve the prediction accuracy by incrementally adding data points to demonstrate recovery actions from the states that cause convergence failures.
As illustrated in \cref{fig:data_augmentation}, in each training iteration $i$, we train an oracle $\mathcal{O}_i$ based on the current dataset $\mathcal{D} = \mathcal{D}_0\cup\mathcal{D}^*_1\cup\ldots\cup\mathcal{D}^*_{i-1}$, where $\mathcal{D}_0$ is the initial dataset, and $\mathcal{D}^*_1\cup\ldots\cup\mathcal{D}^*_{i-1}$ are the augmented datasets obtained from previous training iterations. 
The initial dataset $\mathcal{D}_0$ is achieved by rolling out the traditional \gls{rhp} with a 3-step Prediction Horizon (\gls{ph})\footnote{We choose 3-step Prediction Horizon (\gls{ph}), as we find a longer lookahead does not improve the quality of the motion (cost), while increasing the computation time.} over a set of randomly sampled environments, and then extracting the datapoints from the Execution Horizon (\gls{eh}) of each cycle. 
For computing the augmented dataset $\mathcal{D}^*_i$, we firstly use locally-guided \gls{rhp} with the currently trained oracle $\mathcal{O}_i$ to plan trajectories in a \gls{rhp} fashion on the previously sampled environments. 
Then, we use the traditional \gls{rhp} to compute recovery actions, which starts from 1 to 3 cycles before the locally-guided \gls{rhp} fails, until the cycle converges back to the ground-truth trajectory (roll-out of traditional \gls{rhp} on the same terrain). 
We compute the recovery actions from 1 to 3 cycles before the convergence failures, since we observe that the roll-out of locally-guided \gls{rhp} often exhibits large deviations from the ground-truth trajectory in these cycles.
We repeat the process until there is no further improvement on the convergence rate. 

\subsection{Interfacing to the Short-horizon \gls{to}}
\label{sec:interface_to_to_oracle}

To guide the locally-guided \gls{rhp}, we adapt the short-horizon version of \eqref{eq:general_formulation} that only plans the Execution Horizon (\gls{eh}) with the following changes. 
First, we replace the terminal cost with $(\bm{x}_T - \bm{x}^*)^\trsp(\bm{x}_T - \bm{x}^*) + (\bm{p}^1 - \bm{p}^*)^\trsp(\bm{p}^1 - \bm{p}^*)$ that encourages the terminal state $\bm{x}_T$ and the contact location $\bm{p}^1$ to approach the predicted targets $\bm{x}^*$ and $\bm{p}^*$. 
Second, we introduce constraints to narrow down the search space of phase switching timings $t^i$ around their predicted values $(1-\epsilon)\tilde{t}^q \leq t^q \leq (1+\epsilon)\tilde{t}^q$, where $\epsilon$ is a user-defined slack. 
This is because we empirically find that reducing the search space of the phase switching timings can result in more efficient computation than using cost terms to bias their decisions.

\section{Simulation Studies}
\label{sec:result}

In this section, we evaluate the computation performance of multi-fidelity \gls{rhp} and locally-guided \gls{rhp} over a set of multi-contact scenarios in simulation. 
To the best of our knowledge, there is no accessible method focused on accelerating the computation speed of multi-contact \gls{rhp} by simplifying the value function approximation. 
Therefore, to highlight the computation advantage of our methods, we compare them against the traditional \gls{rhp} (baseline) which approximates the value function by planning the prediction horizon with an accurate dynamics model. The video of our simulations can be found at \url{https://youtu.be/STBYJl7jvsg}.

\begin{table*}[ht!]
  \centering
  \caption{Computation performance for the moderate slope terrain (CS1). We use the incremental training scheme presented in \cref{sec:oracle_function} to train the oracle for \gls{lg-rhp}. The training process lasts for 5 training iterations, after which we do not observe any improvements in the convergence rate (see \cref{sec:incremental_training_result}). }
  \label{tab:computation_cs1}
  \begin{tabularx}{\linewidth}{{M{0.08\linewidth}  M{0.07\linewidth} M{0.00\linewidth} M{0.06\linewidth} M{0.06\linewidth} M{0.06\linewidth} M{0.1\linewidth} M{0.0\linewidth} M{0.06\linewidth} M{0.06\linewidth} M{0.06\linewidth} M{0.1\linewidth} }}
    \toprule
    \multicolumn{2}{c}{\multirow{2}{*}{Method}} & & \multicolumn{4}{c}{Episodic Success Rate } & & \multicolumn{4}{c}{Cycle-wise Success Rate }
    \\ \cline{4-7} \cline{9-12}
    & & &  Success (Offline) & Success (Online) & Time Out & Fail to Converge  & & Success (Offline) & Success (Online) & Time Out & Fail to Converge 
    \\
    
    \midrule
    Baseline & 1-Step \gls{ph}  &  & \colorbox{green}{100.0\%} & 0.0\% & 100.0\% & 0.0\% & & 100.0\% & \colorbox{pink}{51.31\%} & 48.69\% & 0.0\% \\
                              
    \hline
                              
    \gls{mf-rhp} 1 (\gls{com}) & \shortstack{1 to 3-Step \\ \gls{ph}}  &  & \colorbox{pink}{0.0\%} & - & - & - & & - & - & - & - \\
    
    \hline
    
    \multirow{3}{*}{\shortstack{\gls{mf-rhp} 2 \\ (Rectangle)}} & 1-Step \gls{ph}  &  & \colorbox{lime!50}{72.50\%} & 0.0\% & 72.50\% & 27.50\% & & 98.83\% & \colorbox{lime!50}{69.22\%} & 29.61\% & 1.17\%  \\
                                    & 2-Step \gls{ph}  &  & \colorbox{lime!50}{76.32\%} & 0.0\% & 76.32\% & 23.68\%  &  & 99.05\% & \colorbox{pink}{44.88\%} & 54.17\% & 0.95\% \\
                                    & 3-Step \gls{ph}  &  & \colorbox{green}{97.37\%} & 0.0\% & 97.37\% & 2.63\% & & 99.91\% & \colorbox{pink}{5.66\%} & 94.25\% & 0.09\% \\
   
    \hline
    
    \multirow{3}{*}{\shortstack{\gls{mf-rhp} 3 \\ (Point)}} & 1-Step \gls{ph}  &  & \colorbox{lime!50}{79.49\%} & 0.0\% & 79.49\% & 20.51\% & & 99.16\% & \colorbox{lime!50}{75.11\%} & 24.90\% & 0.84\% \\
                                        & 2-Step \gls{ph}  &  & \colorbox{lime!50}{83.78\%} & 0.0\% & 83.78\% & 16.22\% & & 99.38\% & \colorbox{lime!50}{67.08\%} & 32.30\% & 0.62\% \\
                                        & 3-Step \gls{ph}  &  & \colorbox{green}{97.5\%} & 0.0\% & 97.5\% & 2.5\% & & 99.91\% & \colorbox{pink}{49.78\%} & 50.14\% & 0.09\% \\
    
    \hline
    
    \gls{lg-rhp} & -  &  & \colorbox{lime!50}{75.68\%} & \colorbox{green}{67.57\%} & 8.11\% & 24.32\% & & 99.05\% & \colorbox{green}{98.63\%} & 0.42\% & 0.95\%\\
    \hline
    \bottomrule
  \end{tabularx}
  \vspace{-3mm}
\end{table*}

\subsection{Evaluation Setup}
We consider the following two types of terrains: 1) moderate slope terrain (\cref{fig:snapshots_rubbles}) and 2) large slope terrain (\cref{fig:largeslope_snapshots}). 
Planning multi-contact motions on these terrains can be challenging. 
The key issue is that the admissible contact force is limited by the orientation of the surface in contact. 
As a result, in order to find a feasible momentum trajectory of the \gls{com}, the planning algorithm has to carefully select the contact locations and the timings~\cite{ponton2021efficient}. 

On these terrains, we use each \gls{rhp} framework to offline compute centroidal trajectories of the humanoid robot Talos~\cite{stasse2017talos} in a receding horizon fashion. 
To give more detail, we consider a \gls{rhp} loop where each planning cycle aims to compute the motion plan to be executed for the next cycle. 
Under the assumption that the controller can track the planned motion without having large deviations, we enforce the motion plan of the next cycle to always starts from the terminal state of the Execution Horizon (EH) planned for the current cycle. 

To highlight computation benefit of our proposed \gls{rhp} frameworks, we consider an online setting, where we impose computation time limit in each cycle. 
To give more detail, we denote a cycle achieves online computation, if the \gls{to} converges within the time budget---the duration of the motion to be executed (\gls{eh}) in the current cycle. 
In the case of the \gls{to} fail to converge within the time budget, we still leave the \gls{to} to compute until convergence, unless there is no solutions found (fail to converge). 

We test all the \gls{rhp} frameworks on the terrains that are unseen during the training of the locally-guided \gls{rhp}, and we refer to the trial on each terrain as an episode. 
To validate the dynamic feasibility of the planned trajectories, we track them by using a whole-body inverse dynamics controller~\cite{del2016robustness} in simulation. 
More specifically, in our simulation, we employ the inverse dynamics controller to verify the existence of feasible contact forces and joint torques required for executing the planned motions. This verification is conducted within a forward integration loop implemented based on Pinocchio~\cite{carpentier2019pinocchio}.

\subsection{Implementation Details}
\label{sec:implementation_details}

We use the software package CasADi~\cite{Andersson2018} to model the \gls{to} problems in Python, and solve them using the interior-point method of KNITRO 10.30~\cite{byrd2006knitro}. 
Furthermore, we also leverage CasADi to provide gradients and Hessians via automatic differentiation. 
Although analytical methods such as Pinocchio~\cite{carpentier2019pinocchio} can further improve the computation speed of these derivatives, we use CasADi since computing the derivatives with automatic differentiation is not the major computation bottleneck (only 10\% of the total computation time is used for derivative computation). 
In our work, all the computations are carried out on a desktop with an Intel i9-CPU (3.6GHz) and 64GB memory.

For locally-guided \gls{rhp}, we train separate oracles for the two types of the terrains. 
This is because we find the data distributions of these terrains have different modalities, i.e. when traversing the large slopes, the robot tends to exhibit larger momentum variations than walking on the moderate slopes. 
Mixing these data points together can lead to a discontinuous and unbalanced dataset, on which a single neural network model can struggle to interpolate. 
In \cref{sec:discussion}, we discuss the potential options that can generalize across these two modalities. 
To train the oracle, we use the incremental training scheme described in \cref{sec:oracle_function}, and we show that this training scheme can improve the prediction accuracy of the oracle in \cref{sec:incremental_training_result}. 
We train the oracle with the ADAM algorithm~\cite{KingBa15}, and we set the batch size to 1280 datapoints and the learning rate to $1 \times 10^{-5}$. We use the mean square error function as the loss function.
When testing the locally-guided \gls{rhp}, we employ the oracle with the best prediction accuracy achieved after 5 training iterations.

We generate the training environments and testing environments by a random sampling process. 
Specifically, for a given set of contact surfaces, we first determine their orientations with uniform sampling, i.e., rotating either around the roll or the pitch axis. 
After that, we use the uniformly sampling to decide the slope angle of each contact surface. 
To increase the chance of covering a large variety of training examples, we sample 3852 moderate slope terrains and 6245 large slope terrains for training. 

\subsection{Case Study 1 (CS1): Moderate Slope}
\label{sec:cs1_result}

\begin{figure*}[t]
    \centering
    \def\svgwidth{0.9\linewidth}
        \input{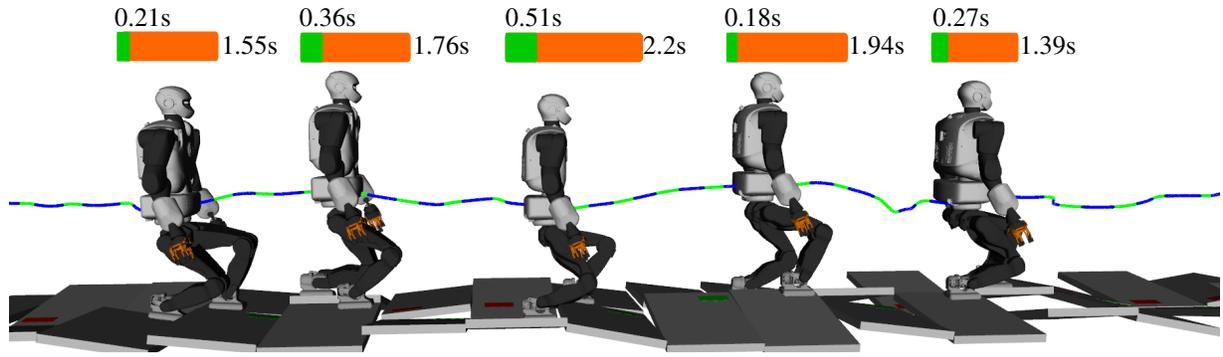}
        \vspace{-2mm}
\caption{Simulation result of locally-guided \gls{rhp} on moderate slopes (5-12 degrees). Owing the shortened planning horizon, our locally-guided \gls{rhp} achieves the fastest computation speed, which allows online \gls{rhp} for the entire episode, i.e., the computation time for planning the next cycle (green bar) is always smaller than the motion duration for the current cycle (orange bar) in the episode shown in this figure.
The video is available at \url{https://youtu.be/STBYJl7jvsg}.}
\label{fig:snapshots_rubbles}
\end{figure*}

In this section, we present the experiment result of our first case study: walking on the moderate slopes (\cref{fig:snapshots_rubbles}). 
Although we can quickly find quasi-static motions for this type of terrain\footnote{This is because the force vectors from the friction cone associated to each contact surface can cancel the gravity.}~\cite{song2021solving}, we are interested in planning dynamic walking motions using \gls{to}. 
This provides us with a unified approach to handle non-quasi-static cases, such as the large slope terrain. Furthermore, walking dynamically can also allow more efficient task completion. 
In this case study, we set the slack of the phase switching timing constraints for locally-guided \gls{rhp} as $\epsilon=0.6$. 
We find this can increase the chance of finding a solution (enlarged search space) without sacrificing much computation time. 

We evaluate the performance of the each \gls{rhp} framework based on episodic success rate and cycle-wise success rate in both the offline and online setting. 
We declare an episode is successful if the chosen \gls{rhp} framework can compute the motion plan for all the cycles within the episode. 
In this case study, we define that each episode contains a maximum number of 28 cycles. 

\begin{figure*}[ht]
    \centering
    \def\svgwidth{0.8\linewidth}
        \input{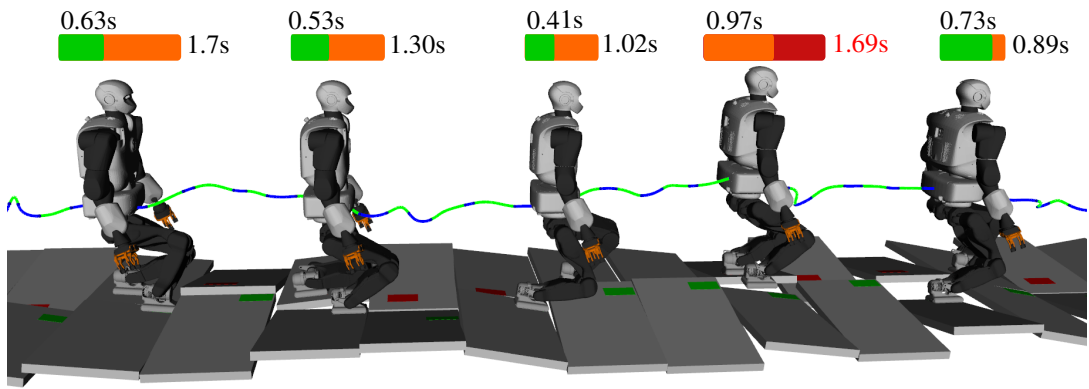}
        \vspace{-2mm}
\caption{Simulation result of our third multi-fidelity \gls{rhp} candidate with 1-step prediction horizon on the moderate slopes (5-12 degrees). Owing to the reduced model complexity in the prediction horizon, our multi-fidelity \gls{rhp} can achieve online computation, e.g., the snapshots where the computation time for planning the next cycle (the green bar) is less than the duration of the motion being executed in the current cycle (the orange bar). Nevertheless, there could be occasions when the computation time exceeds the motion duration of the current cycle (the red bar).
The video is available at \url{https://youtu.be/STBYJl7jvsg}.
}
\label{fig:ponton_single_motion}
\end{figure*}

As the \cref{tab:computation_cs1} indicates, in the offline mode (unlimited time budget), the baseline can achieve a 100\% episodic success rate on the moderate slope terrain with only 1-step Prediction Horizon (\gls{ph}). 
This means the baseline can successfully find solutions for all the cycles (100\% offline cycle-wise success rate). 
However, due to the non-convex nature of the centroidal dynamics constraint, the baseline has nearly half of the cycles (48.69\%) fail to converge online (time out). 

In contrast, the experiment result of multi-fidelity \gls{rhp} demonstrates that we can \textit{improve the computation efficiency by trading off the model accuracy in the Prediction Horizon (\gls{ph}). However, the trade-off cannot be arbitrary.} 
For instance, although our first multi-fidelity \gls{rhp} candidate features the simplest model in the \gls{ph} (linear \gls{com} dynamics), it always fails to complete an episode after a few cycles, no matter how many steps lookahead we assign to the \gls{ph}. 
This suggests that considering the angular dynamics in the \gls{ph} is critical.
Indeed, despite we consider convex relaxed angular dynamics constraints in our second and third multi-fidelity \gls{rhp} candidate, both of them can achieve an offline episodic success rate of 72.5\% to 79.49\% with only 1-step \gls{ph}. 
Furthermore, owing to the relaxed dynamics model employed in the \gls{ph}, our second multi-fidelity \gls{rhp} candidate can achieve 69.22\% of the cycles converging online, which outperforms the baseline (51.31\% cycles computed online). 
This demonstrates that reducing the non-convexity of the \gls{to} problem can improve the computation efficiency. 
Moreover, as our third multi-fidelity \gls{rhp} candidate reduces the dimensionality of the convex relaxation (switching to point foot), it further improves computation efficiency and increases the online cycles-wise success rate to 75.11\%. 
In \cref{tab:average_compute_time_multi_fidelity}, we list the average computation time of the baseline and our second and third multi-fidelity \gls{rhp} candidates.

\begin{table}[t]
  \centering
  \caption{Average computation time of the baseline and the multi-fidelity \gls{rhp} candidates with 1-step \gls{ph} on the moderate slope terrain (CS1)} 
  \label{tab:average_compute_time_multi_fidelity}
  \begin{tabularx}{\linewidth}{M{0.335\linewidth} M{0.275\linewidth} M{0.275\linewidth}}
    \toprule
    Method & Avg. Comput. Time & Avg. Time Budget \\ 
    \midrule
    Baseline          & 2.38 +/- 2.66s & 1.77 +/- 0.33s \\
    \gls{mf-rhp} 2 (Rectangle)    & 1.03 +/- 1.06s & 1.20 +/- 0.40s \\ 
    \gls{mf-rhp} 3 (Point) & \textbf{0.90 +/- 0.81}s & 1.21 +/- 0.38s \\ 
    \bottomrule
  \end{tabularx}
  \vspace{-3mm}
\end{table}

On the other hand, we also notice that our second and third multi-fidelity \gls{rhp} candidate still have the risk to fail to converge, i.e. when considering 1-step \gls{ph}, the second and the third multi-fidelity \gls{rhp} candidate fail during 20.52\% to 27.50\% of the episodes due to convergence issues. 
Although we can improve the convergence rate by extending the length of the \gls{ph}, this can increase the dimensionality of the \gls{to} problem and hinders online computation. 
For instance, when considering 3-step \gls{ph}, both of our second and third multi-fidelity \gls{rhp} can achieve a high episodic success rate (97\%) that is close to the baseline.
Nevertheless, this gives rise to 50.14\% and 94.25\% cycles fail to achieve online computation.
In \cref{fig:ponton_single_motion}, we illustrate a sequence of simulation snapshots of our third multi-fidelity \gls{rhp} candidate with 1-step \gls{ph}, which achieves the best online convergence rate among all multi-fidelity \gls{rhp} candidates. 

Compared to the baseline and multi-fidelity \gls{rhp}, we show that \textit{locally-guided \gls{rhp} achieves the fastest computation speed}, where 98.63\% of the cycles converge online. 
Furthermore, owing to the fast computation, our locally-guided \gls{rhp} can maintain online computation for 67.57\% episodes, whereas the baseline and the multi-fidelity \gls{rhp} struggle to achieve online computation consecutively for a complete episode. 
Additionally, as \cref{tab:time_margin} shows, locally-guided \gls{rhp} only consumes on average 19\% of the time budget. This suggests the potential of using locally-guided \gls{rhp} in real robot control, as the remaining time budget can be allocated to the overheads, e.g. data transmission. 
However, due to the prediction error of the oracle, locally-guided \gls{rhp} also has the chance to fail to converge, i.e. locally-guided \gls{rhp} failed 24.43\% episodes as the robot is directed towards ill-posed states which can cause convergence failures. In \cref{fig:snapshots_rubbles}, we show a sequence of the simulation snapshots for the locally-guided \gls{rhp}. 

\begin{table}[t]
  \centering
  \caption{Average computation time of locally-guided \gls{rhp} v.s. average time budget for the cycles that converged online. } 
  \label{tab:time_margin}
  \begin{tabularx}{\linewidth}{M{0.3\linewidth} M{0.3\linewidth} M{0.28\linewidth}}
    \toprule
    Terrain & Avg. Comput. Time & Avg. Time Budget \\ 
    \midrule
    Moderate Slope (CS1)         & \textbf{0.37 +/- 0.19}s & 1.97 +/- 0.23s \\
    Large Slope (CS2)                     & \textbf{0.36 +/- 0.23}s & 2.01 +/- 0.48s \\ 
    \bottomrule
  \end{tabularx}
  \vspace{-3mm}
\end{table}

\subsection{Case Study 2 (CS2): Large Slope}
\label{sec:cs2_restul}

\begin{table*}[t]
  \centering
  \caption{Computation Performance for the Large Slope Terrain (CS2). We use the incremental training scheme presented in \cref{sec:oracle_function} to train the oracle for \gls{lg-rhp}. The training process lasts for 5 training iterations, after which we do not observe any improvements in the convergence rate (see \cref{sec:incremental_training_result}). }
  \label{tab:computation_cs2}
  \begin{tabularx}{\linewidth}{{M{0.08\linewidth}  M{0.07\linewidth} M{0.00\linewidth} M{0.06\linewidth} M{0.06\linewidth} M{0.06\linewidth} M{0.1\linewidth} M{0.0\linewidth} M{0.06\linewidth} M{0.06\linewidth} M{0.06\linewidth} M{0.1\linewidth} }}
    \toprule
    \multicolumn{2}{c}{\multirow{2}{*}{Method}} & & \multicolumn{4}{c}{Episodic Success Rate } & & \multicolumn{4}{c}{Cycle-wise Success Rate }
    \\ \cline{4-7} \cline{9-12}
    & & &  Success (Offline) & Success (Online) & Time Out & Fail to Converge  & & Success (Offline) & Success (Online) & Time Out & Fail to Converge 
    \\
    
    \midrule
    \multirow{2}{*}{Baseline} & 1-Step \gls{ph}  &  & 78.47\% & 0.25\% & 78.22\% & 21.53\% & & 94.37\% & \colorbox{pink}{22.93\%} & 71.44\% & 5.63\%  \\
                                            & 2-Step \gls{ph}  &  & \colorbox{green}{100.0\%} & 0.0\% & 100.0\% & 0.0\%  &  & 100.0\% & \colorbox{pink}{7.99\%} & 92.01\% & 0.0\% \\

    \hline
                              
    \gls{mf-rhp} 1 (\gls{com}) & \shortstack{1 to 3-Step \\ \gls{ph}}  &  & \colorbox{pink}{0.0\%} & - & - & - & & - & - & - & - \\
    
    \hline
    
    \multirow{3}{*}{\shortstack{\gls{mf-rhp} 2 \\ (Rectangle)}} & 1-Step \gls{ph}  &  & \colorbox{pink}{40.13\%} & 2.36\% & 37.77\% & 59.87\% & & 83.0\% & 43.63\% & 39.37\% & 17.0\%  \\
                                            & 2-Step \gls{ph}  &  & \colorbox{pink}{52.66\%} & 0.27\% &52.39\% & 47.43\%  &  & 89.90\% & 26.58\% & 63.32\% & 10.10\% \\
                                             & 3-Step \gls{ph}  &  & \colorbox{pink}{53.38\%} & 0.0\% & 53.38\% & 46.62\% & & 91.81\% & 8.00\% & 83.80\% & 8.19\% \\
                                            
    \hline
    
    \multirow{3}{*}{\shortstack{\gls{mf-rhp} 3 \\ (Point)}} & 1-Step \gls{ph}  &  & \colorbox{pink}{36.51\%} & 4.89\% & 31.62\% & 63.49\% & & 81.98\% & 52.50\% & 29.48\% & 18.02\% \\
                                            & 2-Step \gls{ph}  &  & \colorbox{pink}{57.27\%} & 0.77\% & 56.50\% & 42.73\% & & 90.92\% & 37.54\% & 53.37\% & 9.08\% \\
                                             & 3-Step \gls{ph}  &  & \colorbox{pink}{58.19\%} & 0.0\% & 58.19\% & 41.81\% & & 92.64\% & 20.65\% & 71.99\% & 7.36\% \\
    
    \hline
    \gls{lg-rhp} & -  &  & \colorbox{lime}{79.9\%} & \colorbox{green}{76.1\%} & 3.8\% & 20.1\% & & 95.99\% & \colorbox{green}{95.0\%} & 0.99\% & 4.01\%\\
    \hline
    \bottomrule
  \end{tabularx}
  \vspace{-3mm}
\end{table*}

In this section, we present the experiment result for the large slope terrain, on which the robot cannot maintain static stability and has to traverse the terrain dynamically. 
We define that each episode starts from the cycle when the large slope is captured inside the lookahead horizon and ends at the cycle when the robot gets off the large slope. For locally-guided \gls{rhp}, we set the slack of the phase switching timing constraints as $\epsilon = 0.15$, as empirically determined to give a good balance between the success rate and the computation speed.

We list the computation performance of each \gls{rhp} frameworks in \cref{tab:computation_cs2}. 
As we can observe, in an offline setting, the baseline can still achieve 100\% episodic success rate on the considered large slope terrains. 
However, this requires the baseline to consider a 2-step \gls{ph}, which can significantly increase the computation complexity.
As a result, the baseline only has 7.99\% of the cycles converging online. 

On the other hand, we find that \textit{multi-fidelity \gls{rhp} candidates struggle to converge for the large slope terrain}. 
To give more detail, similar to CS1, since our first multi-fidelity \gls{rhp} ignores the angular dynamics in the \gls{ph}, it can never complete a single episode on the large slope terrain. 
However, despite our second and third multi-fidelity \gls{rhp} candidate consider convex relaxations of the angular dynamics, they still fail to complete 41.81\% to 63.49\% episodes. 
This result suggests that the convex relaxed models we employed in the \gls{ph} may not be accurate enough to represent the momentum variation of the highly dynamic motion for traversing the large slope, and further investigation on the balance between the model accuracy and computation complexity is needed. 

\begin{figure*}[t]
    \centering
    \def\svgwidth{0.9\linewidth}
        \input{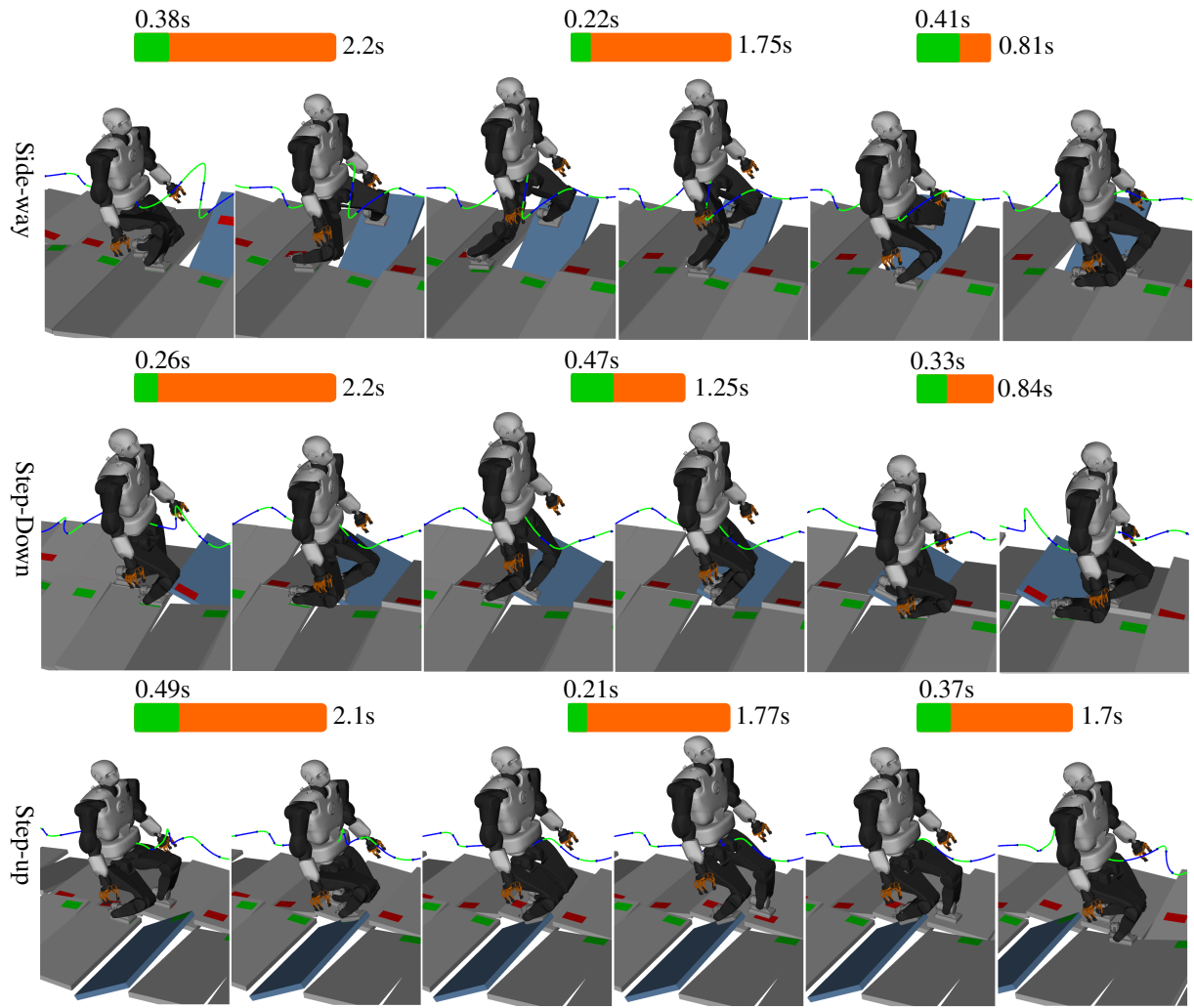}
        \vspace{-2mm}
\caption{Simulation result of locally-guided \gls{rhp} on the large slope terrain (17-25 degrees).
The blue block is the large slope (25 degrees), while the rest are moderate slopes (5-12 degrees). The robot tends to build momentum to achieve dynamic balancing on the large slope. We show that locally-guided \gls{rhp} can be used online, as the computation time of the next cycle (green bar) is smaller than the motion duration of the current cycle (orange bar). The video is available at \url{https://youtu.be/STBYJl7jvsg}.}
\label{fig:largeslope_snapshots}
\end{figure*}

In contrast, despite the increased terrain complexity, \textit{locally-guided \gls{rhp} still achieves the highest computation efficiency among all the \gls{rhp} frameworks}.
More specifically, our experiment result shows that locally-guided \gls{rhp} has 95.99\% cycles successfully converging and 95.0\% of the cycles achieving online computation. 
Owing to the fast computation, our locally-guided \gls{rhp} can also maintain online computation for 76.1\% of the episodes.
For the episodes that fail to achieve online computation consecutively, 3.8\% of them are due to time out, and the rest (20.1\%) are caused by convergence failures. 
Furthermore, as indicated in \cref{tab:time_margin}, our locally-guided \gls{rhp} only consumes on average 18\% of the time budget.
In \cref{fig:largeslope_snapshots}, we show a sequence of simulation snapshots for locally-guided \gls{rhp} on the large slope terrain.

\subsection{Improving Prediction Accuracy with Incremental Training Scheme}
\label{sec:incremental_training_result}
This section demonstrates the effectiveness of our incremental training scheme described in \cref{sec:oracle_function}. 
In \cref{tab:data_augmentation}, we list the episodic success rate of locally-guided \gls{rhp} achieved on the training environments with oracles trained from different iterations of the data augmentation process. Our result shows that adding corrective datapoints of interest can increase the prediction accuracy, which improves the episodic success rate of locally-guided \gls{rhp}. 
We find the success rate saturates after 5 training iterations. 

\begin{table}[t]
  \centering
  \caption{Episodic success rate of different iterations of the incremental training scheme on the training environments.}
  \label{tab:data_augmentation}
  \begin{tabularx}{\linewidth}{M{0.22\linewidth} M{0.1\linewidth} M{0.1\linewidth} M{0.1\linewidth} M{0.1\linewidth} M{0.1\linewidth}}
    \toprule
    Terrain & Iter. 1 & Iter. 2 & Iter. 3 & Iter. 4 & Iter. 5 \\ 
    \midrule
    Moderate (CS1)         & 67.2\%      & 76.2\%  & 80.3\%  & 81.8\%   &82.1\% \\
    Large (CS2)            & 71.5\%      & 75.3\%  & 79.5\%  & 80.4\%   &81.0\% \\
    \bottomrule
  \end{tabularx}
  \vspace{-3mm}
\end{table}

\section{Real-World Experiments}
\label{sec:real_world_exp}

Based on our simulation study (\cref{sec:result}), we  find that our locally-guided \gls{rhp} approach features the best computation efficiency compared to all the \gls{rhp} frameworks considered. 
This computation advantage enables us to demonstrate online receding horizon planning on the torque-controlled humanoid robot platform Talos~\cite{stasse2017talos}. We consider real-world scenarios where online motion adaption is critical, i.e., traversing uneven terrain with unexpected changes. 
Next, we present these robot experiments in detail. We describe the software implementation in \cref{sec:real_world_software_implementation}, and demonstrate the results in \cref{sec:real_world_experiment_result_subsection}. The video of the experiments can be found at \url{https://youtu.be/STBYJl7jvsg}. 

\begin{figure*}[ht!]
    \centering
    \def\svgwidth{\linewidth}
        \input{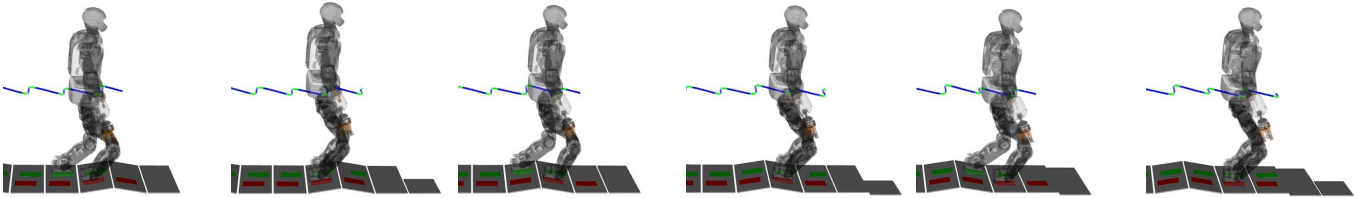}
        \vspace{-7mm}
    \caption{Snapshots for our first real-world experiment in changing environments and the motion planned in each cycle. 
    In this scenario, we change part of the environment from flat surfaces to an up-and-down hill terrain during run-time. 
    We indicate this terrain change by placing a VICON marker next to the terrain, and the planning node  modified the terrain model accordingly, once it detects the VICON marker plate. 
    Owing to the fast computation, our locally-guided \gls{rhp} successfully achieved online receding horizon planning in this scenario, which allows the robot to reliably traverse the terrain. 
    The robot moves from left to right, top to down. The inclination of the slope is 10 degrees. The video is available at \url{https://youtu.be/STBYJl7jvsg}.} 
    \label{fig:changing_env_fig}
\end{figure*}

\subsection{Software Implementation}
\label{sec:real_world_software_implementation}

To achieve the robot experiments, we build a software framework that consists of the following two components: 
1) a planning node, which computes the motion plan in an online receding horizon fashion using locally-guided \gls{rhp} based on the perceived environment, and 2) a robot control stack that executes the planned trajectories in the real-world while considering state feedback of the robot. 

The interplay between these two components are described as following. 
At the beginning of each cycle, the robot control stack  informs the planning node to compute the motion plan for the next cycle, while in the meantime starts executing the motion already planned for the current cycle. 
We assume that robot can always track the planned trajectories without having large deviations. Hence, we define that the motion plan for the next cycle always starts from the terminal state of  the current cycle.
We recall that in each cycle, 
the planning node always computes the Execution Horizon (\gls{eh}) that covers the motion plan of making one step to reach the local objective. 
The prediction of the local objective is based on the preview of the environment. With preview we refer to the current perceived terrain model that is ahead of the robot. 
In our work, we realize the terrain perception based on the VICON motion capture system. 
The terrain perception module identifies different terrain geometries through the detection of a VICON marker plate.
Once the planning node completes the computation, it will send the planned motion to the robot control module for execution in the next cycle. 

To track the planned trajectories, our robot control stack constantly updates the torque command of each joint to achieve the desired motion.
In more detail, in each control loop that runs at 500Hz, the robot control stack firstly decides the desired \gls{com} acceleration and the foot state based on the planned trajectories as well as the state feedback of the robot. 
For instance, the desired \gls{com} acceleration $\ddot{\bm{x}}$ is determined through a PD control law: 
\begin{equation}\label{eq:pd_law}
  \begin{IEEEeqnarraybox}[\IEEEeqnarraystrutmode\IEEEeqnarraystrutsizeadd{2pt}{2pt}][c]{rCl}
    \ddot{\bm{x}} = \bm{K}_p(\bm{x}^{des} - \bm{x}^{fb}) + \bm{K}_d(\dot{\bm{x}}^{des} - \dot{\bm{x}}^{fb}),
  \end{IEEEeqnarraybox}
\end{equation}
where $\bm{K}_p$ and $\bm{K}_d$ are the PD gains, $\bm{x}^{des}$ and $\dot{\bm{x}}^{des}$ are the desired \gls{com} position and velocity interpolated from the planned trajectories, 
and $\bm{x}^{fb}$ and $\dot{\bm{x}}^{fb}$ are the state feedback of the \gls{com} position and velocity. 
To set the desired foot state, we firstly create swing trajectories in between adjacent contacts, and then query the foot state from these swing trajectories for each time step of the control loop. 
After having the desired \gls{com} acceleration and the desired foot state, the robot control stack employs a whole-body inverse dynamics controller developed by PAL Robotics to compute the torque command of each joint. 
To ensure successful tracking, the inverse dynamics controller requires a precise whole-body model to accurately capture the inertial characteristics of the robot. When there are model uncertainties, it is worthwhile to consider robust control strategies to accommodate noise and inaccuracies within the control framework~\cite{del2016robustness}. 

Our software implementation is based on the ROS framework~\cite{ros}, and the communication between the planning node and the robot control stack is achieved through the ROS subscriber/publisher protocol. 
Furthermore, we implement the planning node in Python as described in \cref{sec:implementation_details}, and we develop the robot control stack using C++. 

\subsection{Experiment Result}
\label{sec:real_world_experiment_result_subsection}

\begin{figure*}[ht!]
    \centering
    \def\svgwidth{\linewidth}
        \input{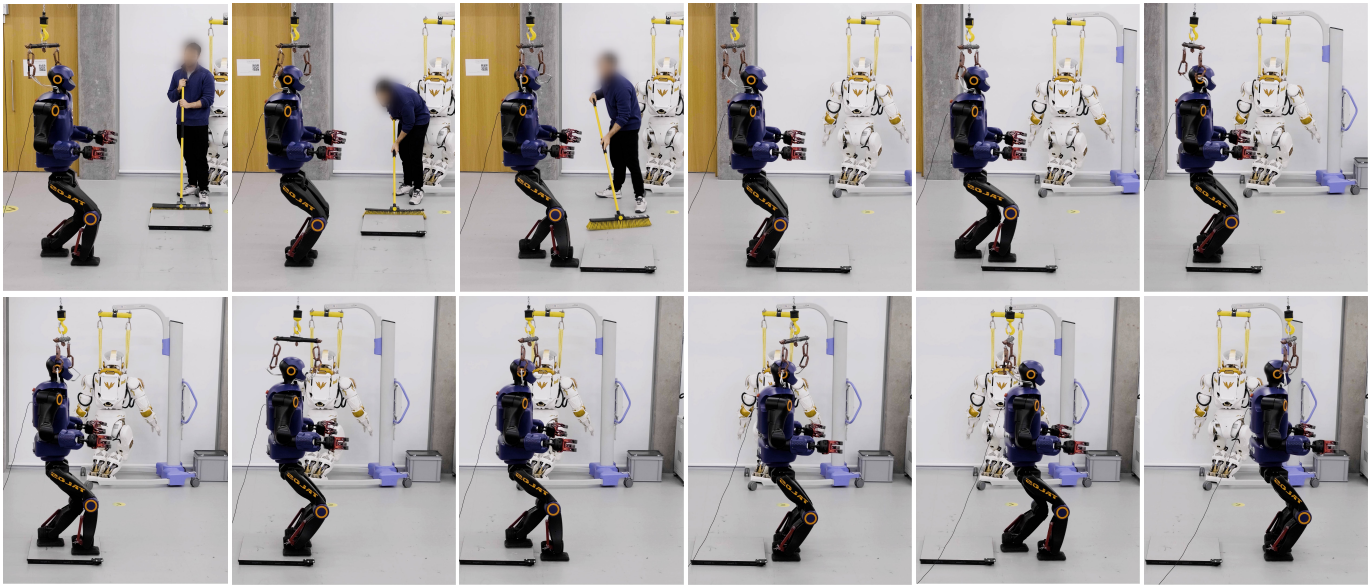}
    \vspace{-7mm}
    \caption{Snapshots of our second real-world experiment in changing environment. In this case, we add a stair (4cm height) while robot is walking. The planning node detects the stair based on the position measurement of the VICON marker plate attached on the stair. 
    In this experiment, our locally-guided \gls{rhp} successfully achieved online computation in each cycle, which allows the robot to safely overcome the stair. The robot moves from left to right, top to down. The video is available at \url{https://youtu.be/STBYJl7jvsg}.} 
    \label{fig:climb_stair_fig}
\end{figure*}

In this section, we present the results of our real-world robot experiments. 
To highlight the benefit of achieving online receding horizon planning, we consider the scenarios where the environment changes during run-time and the robot has to adapt its motion on-the-fly to achieve reliable and continuous operation. 

Specifically, in \cref{fig:changing_env_fig}, we consider a scenario where we change the flat surfaces to an up-and-down hill terrain along the pathway of the robot. 
During the first few cycles, the preview of the environment is considered as flat regions (covered by a curtain). 
While the robot is moving forward,  the flat region changes to an up-and-down hill terrain by removing the curtain. 
The planning node notices the change of the terrain by detecting the VICON marker plate, and updates the environment model accordingly.
During this experiment, locally-guided \gls{rhp} successfully achieved online computation of the contact and motion plans that are consistent to the latest terrain condition perceived the robot. For instance, the average computation time of locally-guided \gls{rhp} is 0.22 +/- 0.076 seconds, which is smaller than the time budget 3.5 seconds. This fast computation speed allows the robot to safely traverse this changing environment. 
In \cref{fig:changing_env_fig}, we show the snapshots of this experiment, as well as the motion plan generated in each cycle along with the terrain model perceived in that cycle. 

In \cref{fig:climb_stair_fig}, we demonstrate another changing environment scenario, where we add a stair during the robot operation. 
Same as in the previous scenario, the locally-guided \gls{rhp} also achieved online receding horizon planning in this scenario. The average computation time is 0.23 +/- 0.1 seconds (the time budget is 3.5 seconds). 
This enables the robot to successfully overcome the newly introduced stair.

\begin{figure*}[ht!]
    \centering
    \def\svgwidth{\linewidth}
\begingroup%
  \makeatletter%
  \providecommand\color[2][]{%
    \errmessage{(Inkscape) Color is used for the text in Inkscape, but the package 'color.sty' is not loaded}%
    \renewcommand\color[2][]{}%
  }%
  \providecommand\transparent[1]{%
    \errmessage{(Inkscape) Transparency is used (non-zero) for the text in Inkscape, but the package 'transparent.sty' is not loaded}%
    \renewcommand\transparent[1]{}%
  }%
  \providecommand\rotatebox[2]{#2}%
  \newcommand*\fsize{\dimexpr\f@size pt\relax}%
  \newcommand*\lineheight[1]{\fontsize{\fsize}{#1\fsize}\selectfont}%
  \ifx\svgwidth\undefined%
    \setlength{\unitlength}{324bp}%
    \ifx\svgscale\undefined%
      \relax%
    \else%
      \setlength{\unitlength}{\unitlength * \real{\svgscale}}%
    \fi%
  \else%
    \setlength{\unitlength}{\svgwidth}%
  \fi%
  \global\let\svgwidth\undefined%
  \global\let\svgscale\undefined%
  \makeatother%
  \begin{picture}(1,0.96805558)%
    \lineheight{1}%
    \setlength\tabcolsep{0pt}%
    \put(0,0){\includegraphics[width=\unitlength,page=1]{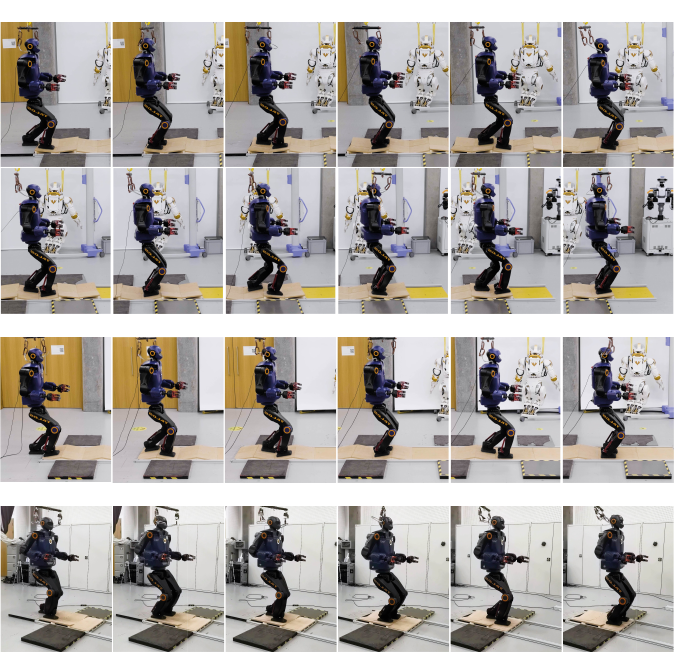}}%
    \put(0.005,0.94276021){\color[rgb]{0.00784314,0.00784314,0.01176471}\makebox(0,0)[lt]{\lineheight{1.25}\smash{\begin{tabular}[t]{l}a) Random Slopes\end{tabular}}}}%
    \put(0.005,0.48){\color[rgb]{0.00784314,0.00784314,0.01176471}\makebox(0,0)[lt]{\lineheight{1.25}\smash{\begin{tabular}[t]{l}b) Up-and-Down Hills\end{tabular}}}}%
    \put(0.005,0.23){\color[rgb]{0.00784314,0.00784314,0.01176471}\makebox(0,0)[lt]{\lineheight{1.25}\smash{\begin{tabular}[t]{l}c) V-shape Terrain\end{tabular}}}}%
  \end{picture}%
\endgroup%

    \vspace{-7mm}
    \caption{Snapshots of our real-world experiment on challenging uneven terrains. Our scenarios include: a) random slopes (the slopes are oriented around either the y-axis or the diagonal axis), b) up-and-down hills, and c) v-shape terrain.  
    The robot moves from left to right, top to down. The inclination of all the slope is 10 degrees. The video is available at \url{https://youtu.be/STBYJl7jvsg}. } 
    \label{fig:uneven_experiments}
\end{figure*}

Furthermore, we also perform real-world experiments on challenging uneven terrains, such as continuously walking on 1) random slopes where the blocks are oriented around either the y-axis or the diagonal axis, 2) up-and-down hill terrain, and 3) the v-shape terrain. The inclination of these slopes are 10 degrees. 
In these experiments, our locally-guided \gls{rhp} achieves online computation for all the cycles. The average computation time is 0.21 +/- 0.06 seconds, and the time budget is 3.5 seconds. 
The snapshots of these experiments are shown in \cref{fig:uneven_experiments}. 

\section{Discussion}\label{sec:discussion}

In this section, we compare the advantages and disadvantages of each \gls{rhp} framework based on our experiment result.

From the result of the baseline, we firstly verify that considering an accurate system dynamic model in the Prediction Horizon (\gls{ph}) 
can guarantee a high convergence rate (100\% for the terrains we considered). 
This is expected as the accurate system dynamics model allows the \gls{ph} to approximate the value function as accurately as possible. 
Furthermore, we also find that although the \gls{ph} does not need to be infinite long, having a \gls{ph} with sufficient length is important to the convergence of the baseline. 
For instance, our experiment result shows that the baseline only requires 1-step \gls{ph} to achieve successful \gls{rhp} on the moderate slope terrain. 
However, to traverse large slope terrain where static stability cannot be maintained, the baseline may need 2-step \gls{ph}. 
Despite the high convergence rate, the downside of the baseline is the long computation time due to the consideration of non-convex centroidal dynamics model, which hinders its online usage. 

To facilitate online multi-contact \gls{rhp}, we explore the trade-off between the computation efficiency and the model accuracy in the \gls{ph}. 
This gives rise to multi-fidelity \gls{rhp}, where we reduce the \gls{to} complexity by employing convex relaxed model in the \gls{ph}. 
From our experiment result, we can draw following conclusions. 
First, we find that the multi-fidelity \gls{rhp} always fails to complete an episode if we only consider linear \gls{com} dynamics in the \gls{ph} (Candidate 1).
This suggests that the convex relaxation employed in the \gls{ph} cannot be arbitrary and considering the angular dynamics is important. This finding leads to our second and third multi-fidelity \gls{rhp} candidates, where we model the angular dynamics with a convex relaxation. 
The results show that our second and third multi-fidelity \gls{rhp} candidates can improve computation efficiency, e.g., for the moderate slope, they outperform the baseline with 20\% to 25\% more cycles computed online (converge within the time budget). 
On the other hand, planning the \gls{ph} with a relaxed model can inevitably affect the accuracy of the value function modeled by the \gls{ph}, which can cause convergence failures. Nevertheless, since the \gls{ph} considers a carefully designed convex relaxation, the occurrence of convergence failures is marginal for our second and third multi-fidelity \gls{rhp} candidates, e.g., only 1\% to 2\% of the cycles fail to converge (see \cref{tab:computation_cs1}).
Although we can further improve the convergence rate of our multi-fidelity \gls{rhp} by extending the length of the \gls{ph}, this increases the dimensionality of the \gls{to} problem, which hinders online computation.
Furthermore, we realize that on the large slope terrain, our multi-fidelity \gls{rhp} fails to complete about half of the episodes, and extending the length of the \gls{ph} does not improve much on the convergence rate. 
This suggests that computing the \gls{ph} with our proposed convex relaxations may lead to inaccurate value function approximations for the large slope terrain. 
We guess the inaccuracy comes from the following two factors. First, the proposed convex relaxations may not be tight enough to capture the momentum changes of highly dynamic motions~\cite{aceituno2018simultaneous}. 
Second, the manually fixed phase switching timings in the \gls{ph} can be invalid for modeling such dynamic motions. 
To conclude, the result of multi-fidelity \gls{rhp} successfully demonstrates that we can achieve online multi-contact \gls{rhp} by relaxing the model accuracy along the planning horizon. 
For future studies, we believe it is worthwhile to improve the performance of multi-fidelity \gls{rhp} in challenging scenarios such as the large slope case, e.g., finding tight convex relaxation of the dynamics and convex formulation of the contact timing optimization.

To further improve the computation efficiency of multi-contact \gls{rhp}, we propose locally-guided \gls{rhp} where we approximate the value function with a learned model. 
More specifically, we introduce an oracle to predict local objectives for achieving a given task, and we then construct local value functions to attract the Execution Horizon (\gls{eh}) towards these predicted local objectives. 
This approach features a shortened planning horizon (only plans the the \gls{eh}) and we demonstrate that locally-guided \gls{rhp} can achieve the best online convergence rate in simulation (95\% to 98.63\% cycles converged online) compared to the baseline and the multi-fidelity \gls{rhp}. 
This computation advantage also enables us to demonstrate online receding horizon on our real-world humanoid robot platform Talos in dynamically changing environments (\cref{sec:real_world_exp}).
However, locally-guided \gls{rhp} still struggles in the following two cases.
First, the oracle can have prediction errors due to imperfect fitting and insufficient data coverage. 
This can lead to inaccurate value functions which direct the robot towards ill-posed states and cause convergence failures. 
Although we can mitigate this issue by an incremental training scheme which demonstrates recovery actions from unseen states, we find it is hard to cover all the possible combinations of the robot state and environment models.
To further improve the prediction accuracy, we believe it is worthwhile to enhance the sampling methods to cover the input space of the oracle more effectively. 
Meanwhile, exploring methods to impose safety constraints in the short-horizon \gls{to} could be also beneficial for improving the convergence rate. 
Second, although locally-guided \gls{rhp} only computes the \gls{eh}, it is still a nonlinear programming problem that has no guarantee on computation time and can fail to convergence online. 
To alleviate this issue, a viable option is to reduce the number of decision variables by representing trajectories with parameterized curvatures, e.g. Bézier Curves~\cite{fernbach2020c}.
Moreover, as mentioned in Section~\ref{sec:implementation_details}, we find that the datapoints for the two types of terrains exhibit different modalities. 
This can impose challenges when training a single Neural Network on the combined dataset. 
Although we capture the two modalities by using separate Neural Networks, it is worthwhile to explore a more unified approach that can handle multimodal data, e.g. using mixture density networks~\cite{Bishop1994}. 

In this work, we assume the sequence of contact surfaces is predefined~\cite{deits2014footstep,song2021solving} and the selection of gait patterns is given, i.e. the sequence in which the feet make and break contacts with the environment~\cite{gaitselection20iros}. 
Ideally, these discrete decisions should be automatically resolved by the optimization. However, this gives rise to combinatorial problems which are difficult to solve. 
In the future, we suggest to extend both multi-fidelity \gls{rhp} and locally-guided \gls{rhp} to consider the combinatorial aspect of the multi-contact planning problem. 
Furthermore, it is also worthwhile to extend our methods to generalize across different tasks, for example, considering different goal positions and behaviour modes (making a turn and side walking).

\section{Conclusion}\label{sec:conclusion}

In this article, we propose multi-fidelity \gls{rhp} and locally-guided \gls{rhp}, two novel methods that can achieve online multi-contact \gls{rhp} on uneven terrains. 
The core idea of our methods is to find computationally efficient approximations of the value function. 
To this end, multi-fidelity \gls{rhp} approximates the value function by computing the prediction horizon with convex relaxed models. 
Alternatively, locally-guided \gls{rhp} focuses on learning a value function model, in which we train an oracle to predict local objectives for completing a given task, and we then build local value functions based on these local objectives. 

The experiment result of the multi-fidelity \gls{rhp} demonstrates that it is possible to achieve online computation by relaxing the model accuracy in the prediction horizon. 
This approach is straight-forward to implement.
However, considering relaxed models in the prediction horizon can downgrade the accuracy of the value function approximation, which may cause convergence failures.
To improve the performance of multi-fidelity \gls{rhp}, we believe future investigations on the balance between the computation efficiency and the model accuracy is important.

Owing to the shortened planning horizon, our locally-guided \gls{rhp} achieves the best online convergence rate among all the \gls{rhp} frameworks. This computation advantage enables us to demonstrate online receding horizon planning on our real-world humanoid robot platform Talos in dynamically changing environments. 
Nevertheless, we find that the oracle can have prediction errors due to inadequate data coverage and lead to convergence failures.
To alleviate this issue, we employ an incremental training scheme to add datapoints from the states that cause convergence failures. We still found it was hard to achieve 100\% prediction accuracy with this approach, showing that further investigations on improving the learning accuracy is necessary.

\section*{Acknowledgment}
This research is supported by EU H2020 project
Enhancing Healthcare with Assistive Robotic Mobile Manipulation (HARMONY, 101017008) and The Alan Turing Institute. The authors would like to thank Theodoros Stouraitis, Iordanis Chatzinikolaidis, Chris Mower, Jo\~{a}o Moura, Carlos Mastalli for their discussion and feedback of the draft, Pierre Fernbach and PAL Robotics for sharing us with the experience of the robot, Huixin Luo for the help of multi-media editing, and Douglas Howie for setting up the experimental terrain. Moreover, the authors would like to express their sincere gratitude to Andreas Christou, Marina Aoyama, Namiko Saito and Ran Long for their help on running the real-world experiment. 

\ifCLASSOPTIONcaptionsoff
  \newpage
\fi

\bibliographystyle{IEEEtran}
\bibliography{reference}  


\vspace{-7mm}

\begin{IEEEbiography}[{\includegraphics[width=1in,height=1.25in,clip,keepaspectratio]{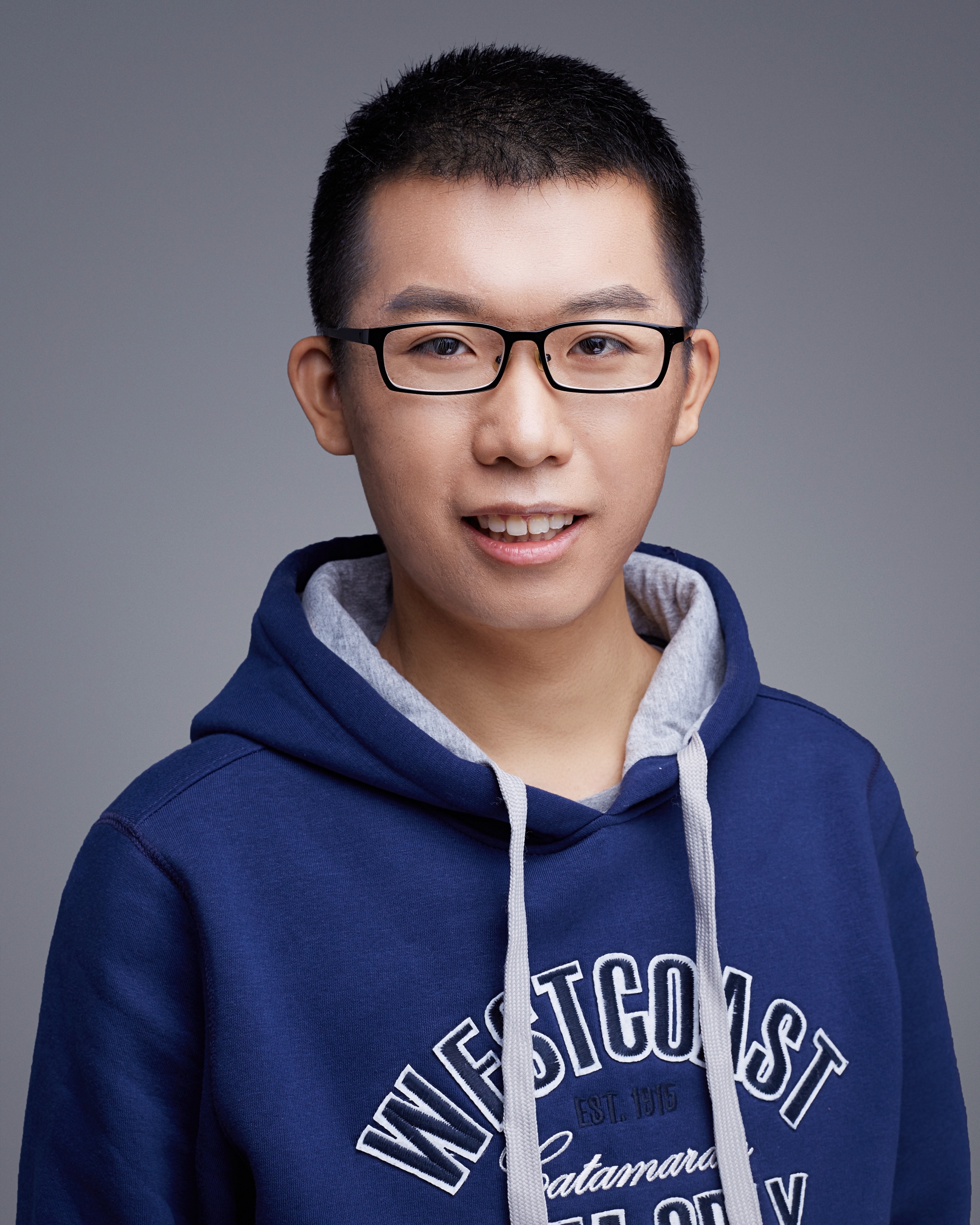}}]{Jiayi Wang} (Member, IEEE) received his Ph.D. degree in robotics from The University of Edinburgh in 2023. He is currently a Research Associate in the School of Informatics at The University of Edinburgh, and affiliated with The Alan Turing Institute. He was a research intern at the Italian Institute of Technology in 2017. Prior to that, he obtained M.Sc. degrees from The University of Edinburgh and The Chinese University of Hong Kong in 2016 and 2015.  
His research interests include humanoid locomotion, multi-contact motion planning, and trajectory optimization. 
\end{IEEEbiography}

\vspace{-7mm}

\begin{IEEEbiography}[{\includegraphics[width=1in,height=1.25in,clip,keepaspectratio]{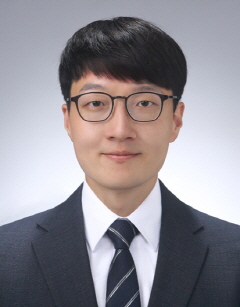}}]{Sanghyun Kim} (Member, IEEE) received a B.S. degree and a Ph.D. degree in robotics from Seoul National University, South Korea, in 2012 and 2020, respectively. He was a postdoctoral researcher at the University of Edinburgh, U.K. in 2020, and a senior researcher at the Korea Institute of Machinery and Materials, South Korea, from 2020 to 2023. He is currently an assistant professor of the department of mechanical engineering at Kyung Hee University, South Korea. His main research interests include the optimal control of mobile manipulators and humanoids.
\end{IEEEbiography}

\vspace{-7mm}

\begin{IEEEbiography}[{\includegraphics[width=1in,height=1.25in,clip,keepaspectratio]{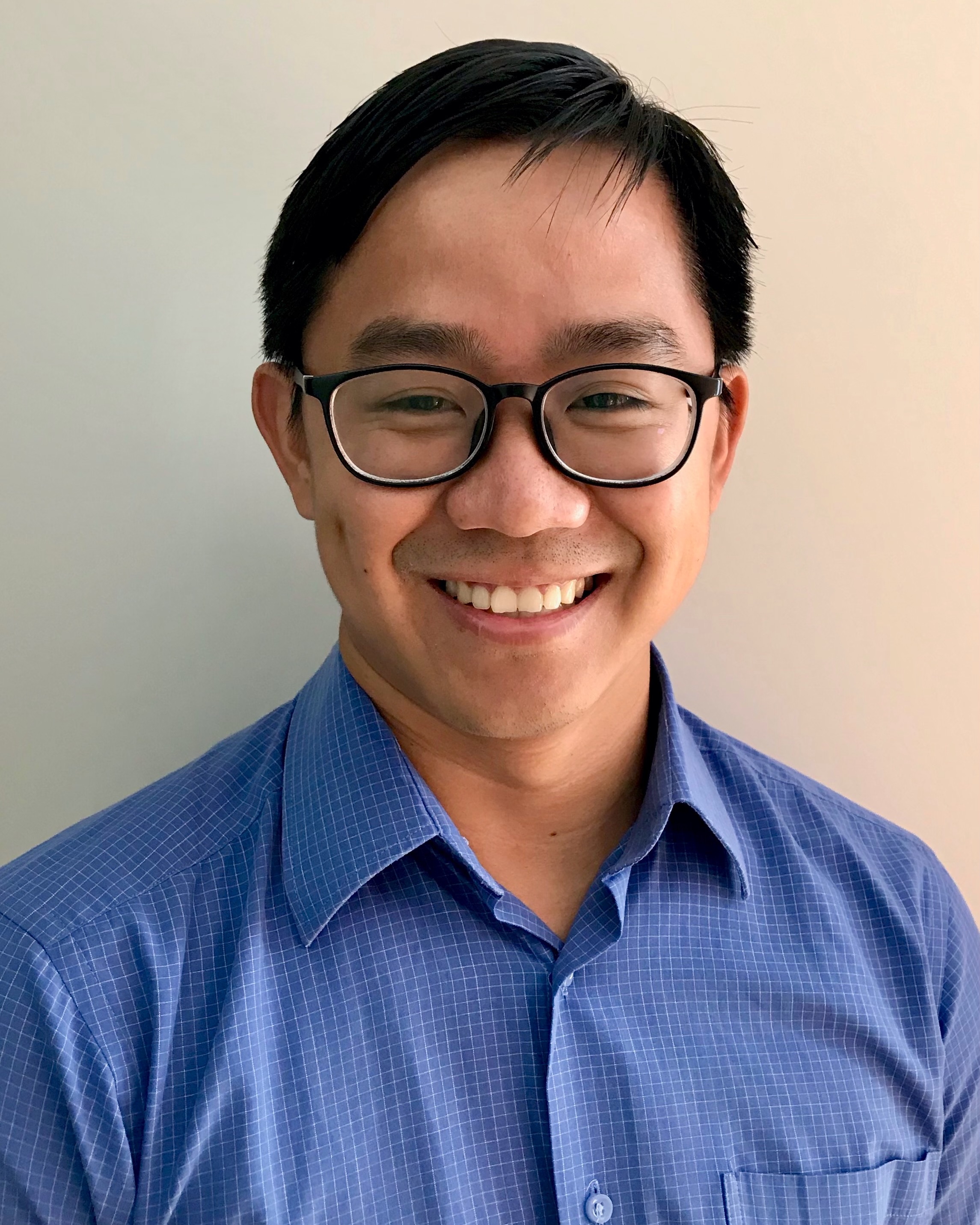}}]{Teguh Santoso Lembono} (Member, IEEE) received the Ph.D. degree in robotics from the École Polytechnique Fedérale de Lausanne (EPFL), Lausanne, Switzerland, in 2022.
He is currently working as an Applied Scientist in Amazon Robotics, Berlin, Germany. He obtained his B. Eng and M. Sc in Mechanical Engineering from NTU and NUS, Singapore in 2012 and 2016, respectively. His research interests include motion planning, control, optimization, and probabilistic machine learning.
\end{IEEEbiography}

\vspace{-7mm}

\begin{IEEEbiography}[{\includegraphics[width=1in,height=1.25in,clip,keepaspectratio]{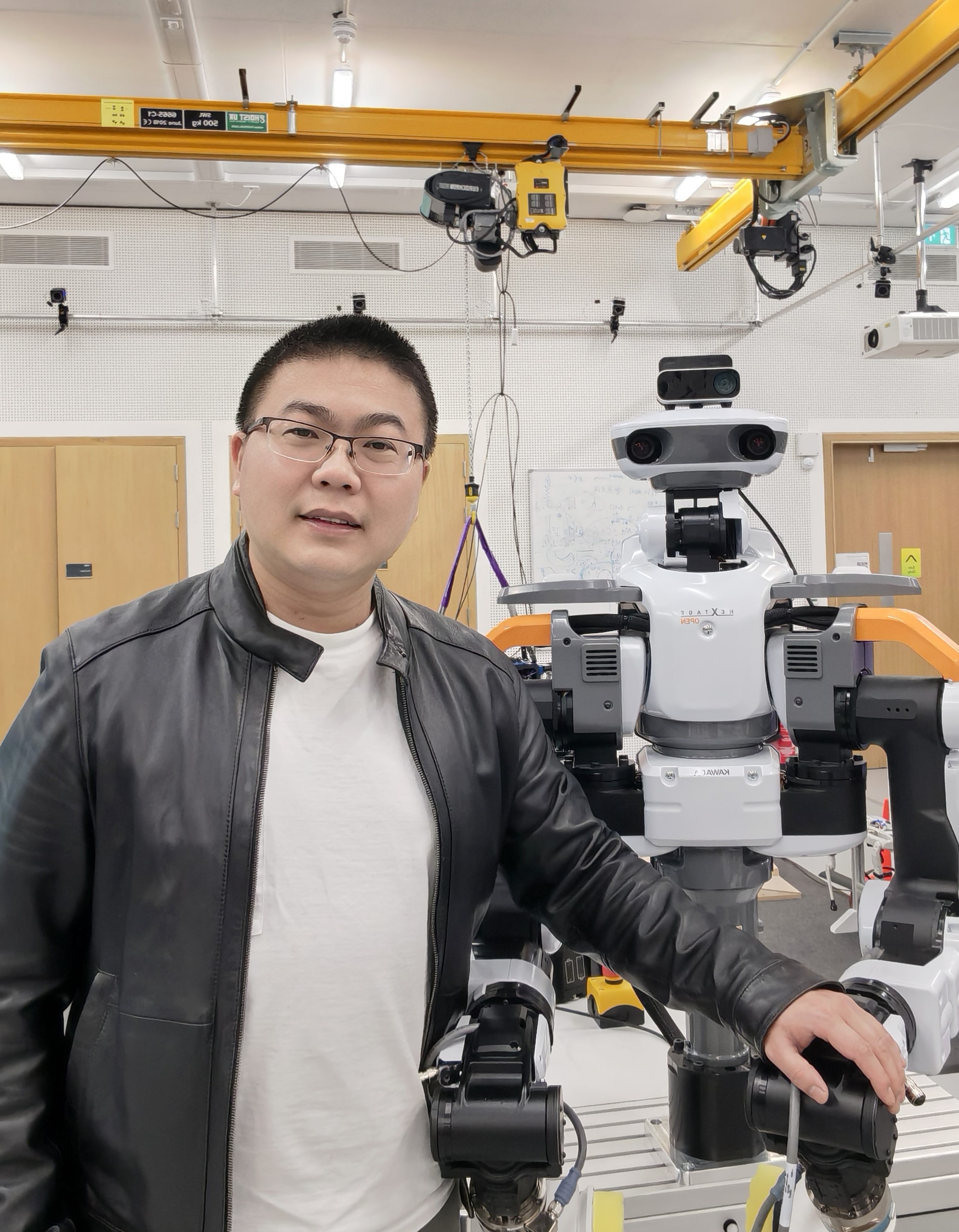}}]{Wenqian Du} received his Ph.D. degree from Institut des Systèmes Intelligents et de Robotique (ISIR) of Sorbonne University, France, in 2021. Then he served as a Huashan-Scholar associate professor at Xidian University, China. He is currently a research associate at the University of Edinburgh, United Kingdom. His research interests include whole-body motion generation of humanoid and quadruped robots, locomotion and balance control, dual-arm mobile manipulation, model-based predictive control, and reinforcement learning.
\end{IEEEbiography}

\vspace{-7mm}

\begin{IEEEbiography}[{\includegraphics[width=1in,height=1.25in,clip,keepaspectratio]{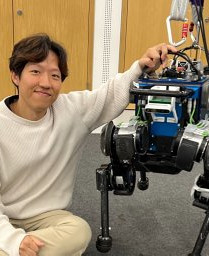}}]{Jaehyun Shim} received a B.Eng. degree in mechanical engineering from the University of Tokyo, Tokyo, Japan in 2015 and an M.A.Sc. degree in mechanical engineering from the University of British Columbia, Vancouver, BC, Canada, in 2018.
He is currently a Software Engineer at the University of Edinburgh. He worked on developing autonomous delivery robots as a Software Engineer at ROBOTIS, Seoul, South Korea. His research interests include optimization-based planning and legged robots. 
\end{IEEEbiography}

\vspace{-7mm}

\begin{IEEEbiography}[{\includegraphics[width=1in,height=1.25in,clip,keepaspectratio]{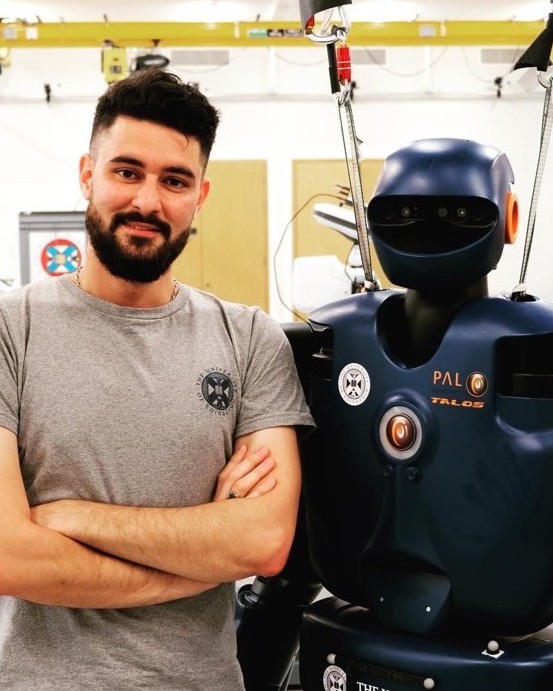}}]{Saeid Samadi} obtained his B.Sc. in mechanical engineering from the University of Tabriz (2016) and M.Sc. from the University of Tehran (2018), followed by a Ph.D. in robotics and AI from CNRS–University of Montpellier (2021), France. He is now a senior researcher at the University of Edinburgh, specializing in humanoid locomotion, loco-manipulation, real-time control of robotic systems, and numerical optimization.
\end{IEEEbiography}

\vspace{-7.5mm}

\begin{IEEEbiography}[{\includegraphics[width=1in,height=1.25in,clip,keepaspectratio]{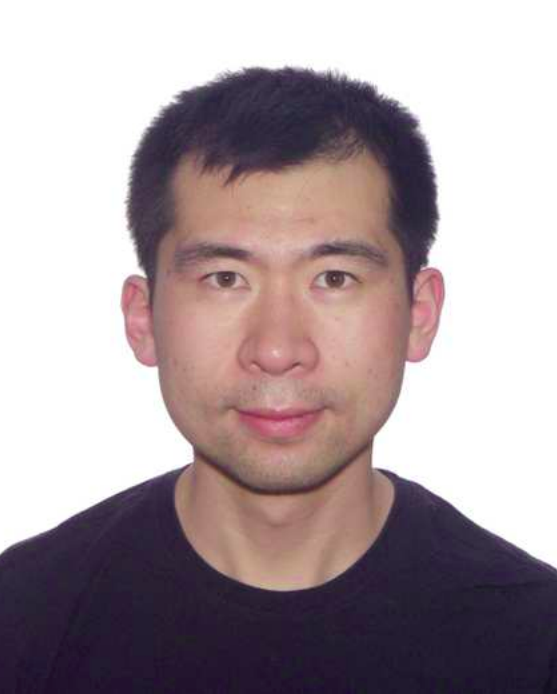}}]{Ke Wang} received the B.E. degree in Automotive Engineering from the Tongji University, Shanghai, China, in 2013, the Master degree in Robotics, Cognition and Intelligence from the Technical University of Munich, Germany in 2016 and the Ph.D. in Robotics from Imperial College London, UK in 2022. He is currently a Senior Robot Research Engineer at Dyson. His research interests include legged robotics, optimization-based control and robot learning.
\end{IEEEbiography}

\vspace{-7.5mm}

\begin{IEEEbiography}[{\includegraphics[width=1in,height=1.25in,clip,keepaspectratio]{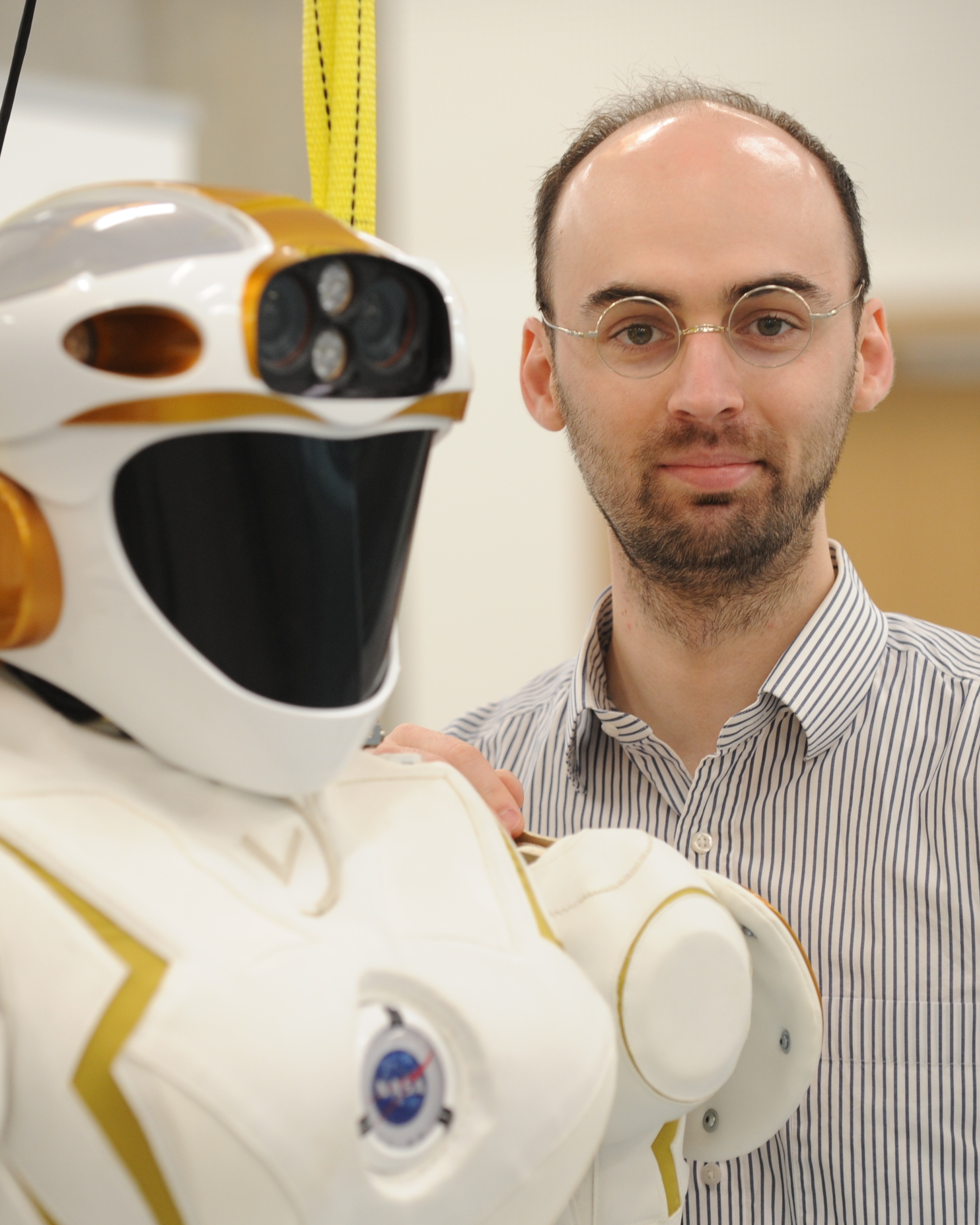}}]{Vladimir Ivan} is VP of Robotics at Touchlab Limited in Edinburgh, working on tactile robots and manipulation. He received the B.Sc. degree in AI and robotics from the University of Bedfordshire, U.K., in 2009, the M.Sc. degree in AI specializing in intelligent robotics, in 2010 and the Ph.D. degree in motion synthesis in topology-based representations from The University of Edinburgh, U.K., in 2014 respectively, where he then worked as a senior researcher until 2022. His research interests include motion planning and modelling, topology, humanoid and legged robotics, manipulation, quantum sensing, and machine learning.
\end{IEEEbiography}

\vspace{-7.5mm}

\begin{IEEEbiography}[{\includegraphics[width=1in,height=1.25in,clip,keepaspectratio]{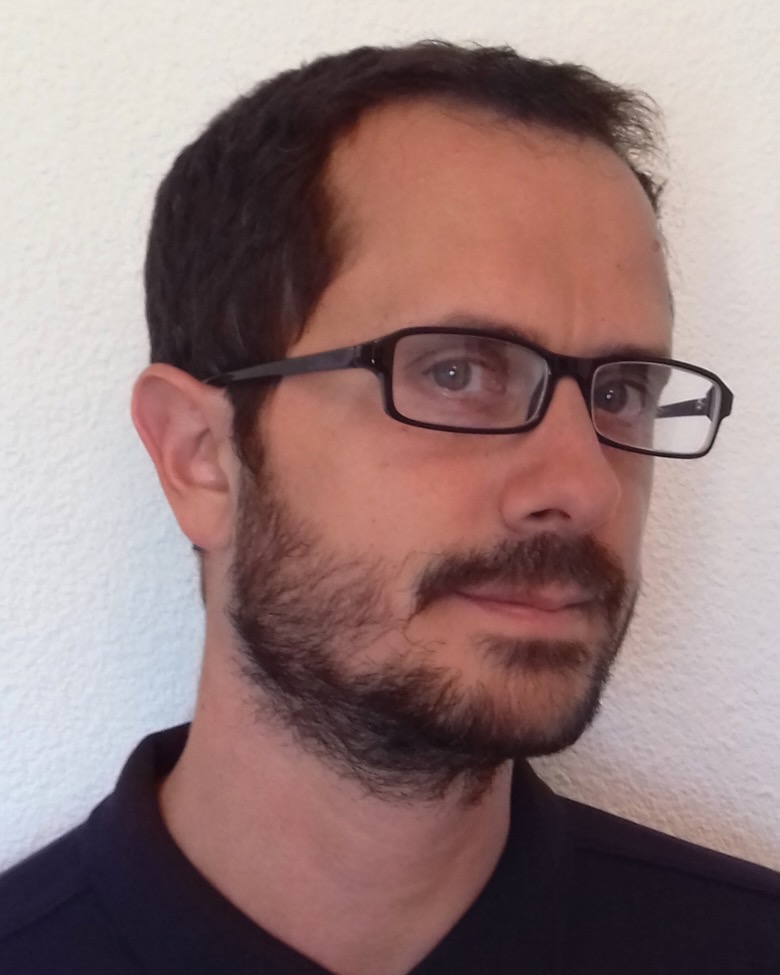}}]{Sylvain Calinon} received the B.Sc. and M.Sc. degrees in microengineering, and the Ph.D. degree in robotics from the {\'E}cole Polytechnique F{\'e}d{\'e}rale de Lausanne (EPFL), in 2001, 2003, and
2007, respectively. He is currently a Senior Research Scientist with the Idiap Research Institute and a Lecturer with EPFL. From 2009 to 2014, he was a Team Leader with the Italian Institute of Technology. From 2007 to 2009, he
was a Postdoc with EPFL. His research interests cover robot learning, human-robot collaboration, optimal control, geometric approaches and model-based optimization. Website: \url{https://calinon.ch}
\end{IEEEbiography}

\vspace{-7.5mm}

\begin{IEEEbiography}[{\includegraphics[width=1in,height=1.25in,clip,keepaspectratio]{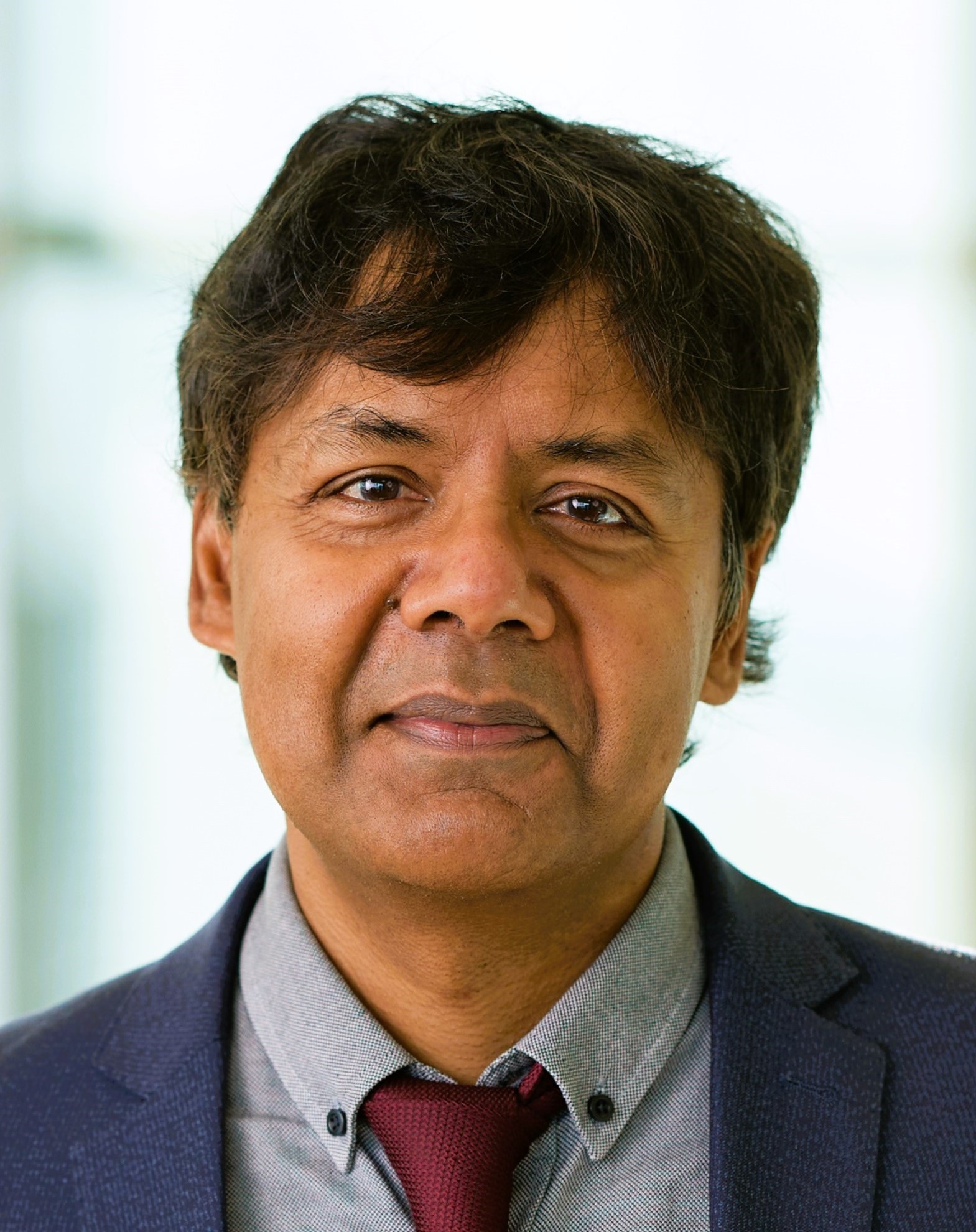}}]{Sethu Vijayakumar} received his Ph.D. in Computer Science and Engineering from the Tokyo Institute of Technology, Japan in 1998. He is the Professor of Robotics at the University of Edinburgh, an adjunct faculty of the University of Southern California, Los Angeles and the founding Director of the Edinburgh Centre for Robotics. His research interests include statistical machine learning, anthropomorphic robotics, multi objective optimisation and optimal control in autonomous systems.
He is the Programme co-Director for Artificial Intelligence at The Alan Turing Institute and 
a Fellow of the Royal Society of Edinburgh.
\end{IEEEbiography}

\vspace{-7.5mm}

\begin{IEEEbiography}[{\includegraphics[width=1in,height=1.25in,clip,keepaspectratio]{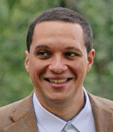}}]{Steve Tonneau} is a lecturer at the University of Edinburgh. He defended his Phd in 2015 after 3 years in the INRIA/IRISA Mimetic research team, and pursued a post-doc in robotics at LAAS-CNRS in Toulouse, within the Gepetto team. His research focuses on motion planning based on the biomechanical analysis of motion invariants. Applications include computer graphics animation as well as robotics. 
\end{IEEEbiography}

\vfill

\end{document}